\documentclass[journal]{IEEEtran}
\usepackage{amsmath,amssymb}
\usepackage{graphicx}
\usepackage{multirow}
\usepackage{booktabs}
\usepackage{xcolor}
\usepackage{hyperref}
\usepackage{colortbl}
\usepackage{cite}
\usepackage{algorithm}
\usepackage{algpseudocode}
\usepackage{pifont}

\title{Unleashing the Potential of All Test Samples: Mean-Shift \\ Guided Test-Time Adaptation}
\author{Jizhou~Han,
Chenhao~Ding,
Songlin~Dong\textsuperscript{\dag},
Xinyuan~Gao,
Qiang~Wang,\\
Yuhang~He,
Yihong Gong, \IEEEmembership{Fellow,~IEEE}
\\

\thanks{$^\dagger$ Songlin~Dong is the corresponding author.}%

\thanks{Jizhou Han, Yuhang He, Qiang Wang and Yihong Gong are with the State Key Laboratory of Human-Machine Hybrid Augmented Intelligence, Institute of Artificial Intelligence and Robotics, Xi'an Jiaotong University, Xi'an 710049, China; Songlin Dong is with the Faculty of Computility Microelectronics, Shenzhen University of Advanced Technology and the Guangdong Provincial Key Laboratory of Computility Microelectronics, Shenzhen 518107, China; Chenhao Ding and Xinyuan Gao are with the College of Software Engineering, Xi'an Jiaotong University, Xi'an 710049, China.}

\thanks{ This article has been accepted for publication in IEEE Transactions on Circuits and Systems for Video Technology. This is the author's version which has not been fully edited and content may change prior to final publication.
Citation information: DOI 10.1109/TCSVT.2026.3676383}%
}

\IEEEpubid{\begin{minipage}{\textwidth}\ \\[30pt] \centering
Copyright \copyright~2026 IEEE. Personal use of this material is permitted. However, permission to use this material for any other purposes must be obtained from the IEEE by sending an email to pubs-permissions@ieee.org.
\end{minipage}}

\begin{document}
\maketitle

\begin{abstract}
Visual-language models (VLMs) like CLIP exhibit strong generalization but struggle with distribution shifts at test time. Existing training-free test-time adaptation (TTA) methods operate strictly within CLIP’s original feature space, relying on high-confidence samples while overlooking the potential of low-confidence ones. We propose MS-TTA, a training-free approach that enhances feature representations beyond CLIP’s space using a single-step k-nearest neighbors (kNN) Mean-Shift. By refining all test samples, MS-TTA improves feature compactness and class separability, leading to more stable adaptation. Additionally, a cache of refined embeddings further enhances inference by providing Mean-Shift-enhanced logits. Extensive evaluations on OOD and Cross-Dataset Benchmarks demonstrate that MS-TTA consistently outperforms state-of-the-art training-free TTA methods, achieving robust adaptation without requiring additional training.
\end{abstract}


\section{Introduction}
\IEEEPARstart{R}{ecent} advancements in visual-language models (VLMs), such as CLIP~\cite{clip} and ALIGN~\cite{ALIGN}, have revolutionized various downstream tasks with their exceptional generalization abilities. These models have demonstrated impressive performance in tasks like image-text matching and zero-shot learning, making them highly effective across a wide range of applications. However, they face significant challenges when there are substantial shifts in the data distribution during testing. As the task distribution evolves, the ability of these models to maintain consistent performance diminishes. This highlights the critical need for methods that allow VLMs, like CLIP, to quickly adapt to new, unseen data distributions in real-world settings.

\begin{figure}[t]
    \centering
    \includegraphics[width=0.9\linewidth]{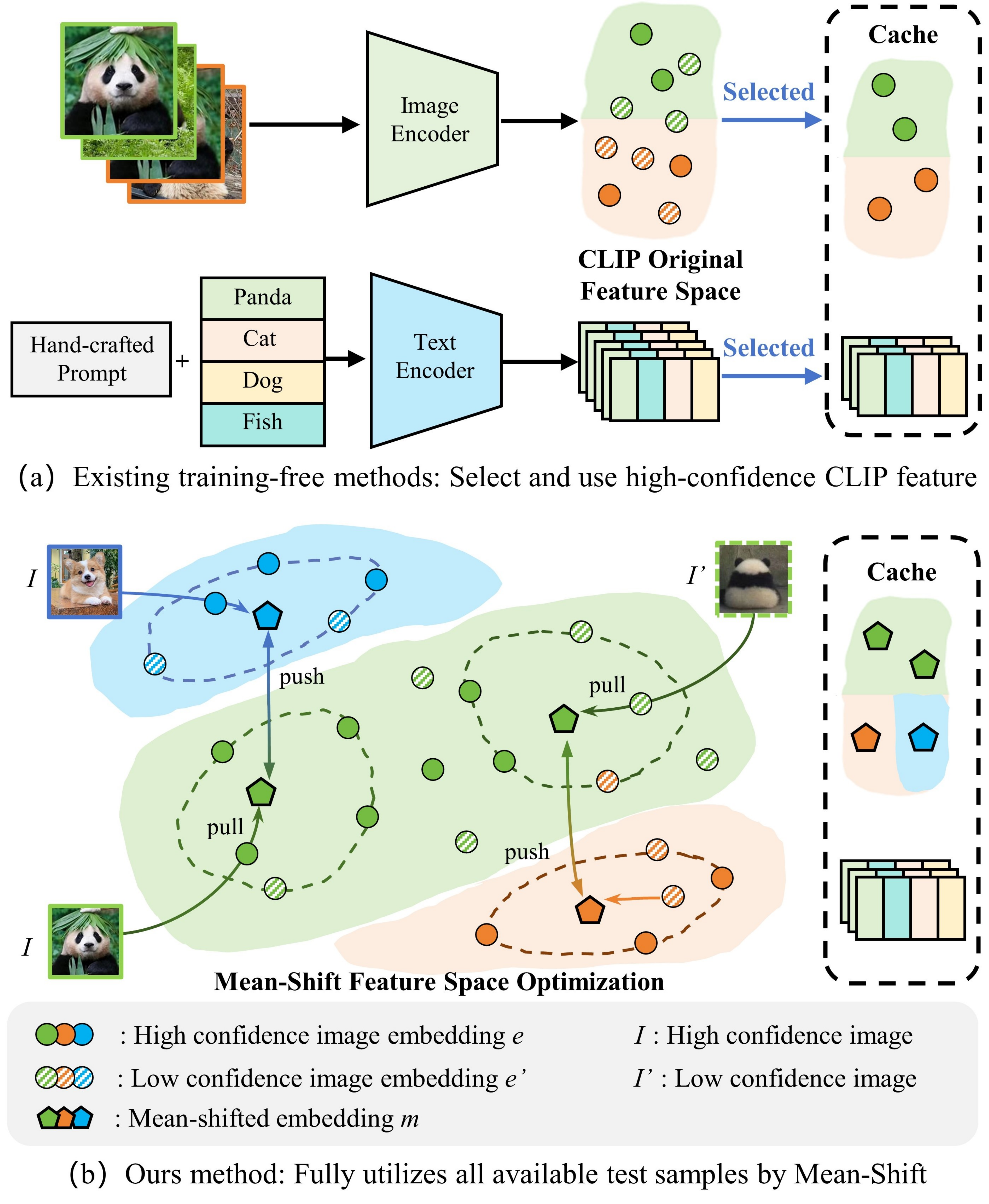}
    \caption{{Illustration of the difference between our method and previous approaches and the proposed  Mean-Shift Guided Test-Time Adaptation.}}
    \label{fig:intro}
\end{figure}

Various TTA approaches have been proposed to address the adaptation challenge. These can be broadly categorized into training-required and training-free methods. Training-required approaches, such as TPT \cite{tpt} and its variants, such as DiffTPT\cite{difftpt} and HisTPT\cite{Zhang2024Historical}, optimize model parameters, including learnable prompts, using self-supervised objectives like entropy minimization. These methods enhance model adaptation but have a significant computational cost, making them impractical for real-time applications. On the other hand, training-free TTA methods leverage feature retrieval and memory-based strategies to modify predictions without updating model parameters. Approaches like Test-Time Adaptation via Dynamic Caching (TDA) \cite{tda} and BoostAdapter \cite{Zhang2024BoostAdapter} employ a dynamic cache to store high-confidence samples, refining predictions via nearest-neighbor retrieval.

However, as illustrated in Fig.~\ref{fig:intro}(a), existing training-free TTA methods operate strictly within CLIP’s original feature space, assuming it is already optimal for adaptation. These methods selectively utilize only ``high-confidence'' samples while overlooking the potential of ``low-confidence'' ones. {In practice, such low-confidence samples often lie near decision boundaries or correspond to rare target-domain patterns; incorporating them when refining the feature space helps shape more accurate decision boundaries and thus improves the generalization ability of TTA models under distribution shifts.} As a result, they heavily rely on the quality of CLIP’s original features. While CLIP’s strong generalization capability benefits these approaches, it also imposes a performance ceiling, limiting their ability to refine and adjust feature representations. This inherent dependence on CLIP’s feature space not only restricts the flexibility of adaptation but also hinders further improvement beyond the model’s initial generalization ability. This raises two questions:

\noindent\textit{\textbf{Can CLIP’s original feature space be further optimized for adaptation? Can so-called “low-quality” samples be refined rather than disregarded?}}

Mean-Shift \cite{meanshift,cheng1995mean} is an unsupervised clustering method that iteratively shifts features toward dense regions in the data distribution, improving their alignment with underlying structures. By utilizing local neighborhood relationships~\cite{fukunaga1975estimation}, it enhances feature compactness. It guides samples toward more representative cluster centers, relying solely on the intrinsic data distribution rather than explicit labels or high-confidence predictions. {From a test-time adaptation perspective, Mean-Shift is appealing because it can refine feature representations using only unlabeled test data by shifting embeddings toward local density modes, which naturally fits the goal of adapting CLIP under distribution shifts without extra training. In MS-TTA, we therefore adopt a lightweight single-step kNN Mean-Shift instead of the classical iterative variant to keep the computation efficient and stable in the online setting.}

Inspired by the effectiveness of Mean-Shift in unsupervised clustering and feature refinement, we introduce MS-TTA, a novel test-time adaptation framework that enhances feature representations beyond CLIP’s original feature space. {Different from previous methods that implicitly assume the CLIP feature space is inherently optimal and mainly adapt the model using high-confidence pseudo-labeled samples, MS-TTA applies a single-step kNN mean-shift to all test samples so that both high- and low-confidence examples can contribute to adaptation in a parameter-free and unsupervised manner.} As illustrated in Fig.~\ref{fig:intro}(b), MS-TTA not only enhances the quality of individual samples during testing but also allows refined samples to improve others over time. Given a test image, MS-TTA first extracts its feature embedding using CLIP’s {visual encoder} and applies Mean-Shift clustering to refine it based on its nearest neighbors. This process shifts low-quality embeddings toward more reliable feature clusters, improving their discriminability and alignment with high-quality samples. Additionally, previously refined samples contribute to the adaptation of new test samples, further improving intra-class compactness and inter-class separability. To support this process, MS-TTA maintains a cache of refined embeddings, which is used to compute Mean-Shift-enhanced logits during inference. These refined logits are then combined with CLIP’s original predictions, resulting in a more robust classification. By directly integrating feature refinement into test-time adaptation, MS-TTA establishes a self-improving mechanism that progressively enhances the entire feature space, ensuring stability and effectiveness under distribution shifts while remaining entirely training-free.

The key contributions of this work are as follows: We introduce MS-TTA, a training-free test-time adaptation framework that refines all test samples using Mean-Shift, enhancing feature quality beyond CLIP’s original space. By leveraging both high- and low-quality samples, MS-TTA improves feature compactness and class separability, enabling more effective adaptation. Extensive evaluations across OOD and Cross-Dataset Benchmarks show that MS-TTA outperforms state-of-the-art training-free TTA methods, ensuring robust adaptation under distribution shifts.

\section{Related Work}

\subsection{Test-time adaptation (TTA)}
TTA has emerged as a critical area of research for handling distribution shifts at test time without access to training data~\cite{tta1,memo,ptta,domainadaptor,dota,zhao2025expamoe}. 
In contrast to incremental learning~\cite{zenke2017continual,han2025goal,wang2025boostdil,han2025biag}, and semi-supervised paradigms~\cite{vaze2022gcd,wen2023parametric,han2025csua}, TTA typically operates on unlabeled test streams and adapts without additional annotation supervision. Existing TTA methods can be broadly categorized into training-required and training-free approaches.

\textit{Training-required methods} optimize model parameters during test-time to adapt to distribution shifts. For instance, TPT~\cite{tpt} optimizes adaptive text prompts through entropy minimization, leveraging AugMix~\cite{augmix} to generate diverse test image augmentations. DiffTPT~\cite{difftpt} extends this approach by incorporating the Stable Diffusion Model~\cite{stable} to create more varied augmentations and filter them based on cosine similarity to the original image. Similarly, HisTPT~\cite{Zhang2024Historical} leverages historical test data to refine prompts for better adaptation. DPE~\cite{DPE} proposes a dual-prototype evolving paradigm for vision-language models, which maintains visual and textual prototypes together with a memory bank of test features and residual adapters, and iteratively updates these prototypes during testing to improve Cross-Dataset generalization. While these methods demonstrate strong adaptation performance, they rely on backpropagation for prompt optimization, which limits their efficiency in fast adaptation scenarios. 
Beyond image classification, parameter-updating TTA has also been validated across diverse tasks, including leveraging audio cues for video model adaptation~\cite{zeng2025audio_tta}, question-type–aware debiasing for test-time VQA~\cite{liu2024qed_tta}, self-supervised adaptation for personalized gaze estimation~\cite{wu2024ttagaze}, and camera-aware recurrent learning with Earth Mover's distance for person re-identification~\cite{chen2023cameraaware_tta}. {In addition, TT-WA~\cite{liu2025test} propose Meta Batch Normalization for test-time adaptation in real-world video adverse weather restoration, further highlighting the versatility of parameter-updating TTA in low-level video restoration tasks. FSTTA~\cite{FSTTA} instead targets online vision-and-language navigation and decomposes gradients and parameters into fast and slow components, using confidence-guided entropy minimization to coordinate high-frequency fast updates and low-frequency slow updates.} {Compared with these task-specific, parameter-updating frameworks, our work instead focuses on training-free test-time adaptation for CLIP-based single-image classification, where the backbone remains frozen and no gradient updates are performed at test time.}

\textit{Training-free methods} aim to adapt models without updating parameters, making them more efficient for real-time applications. Test-Time Adaptation via Dynamic Caching (TDA)~\cite{tda} introduces a cache model inspired by Tip-Adapter~\cite{tip}, which stores representative test samples and refines predictions by comparing incoming samples with the cache. BoostAdapter~\cite{Zhang2024BoostAdapter} dynamically adjusts feature representations during test time to improve robustness to distribution shifts. {BCA~\cite{BCA} further introduces a Bayesian test-time adaptation framework for vision-language models, which maintains category-level statistics in a memory and continuously updates class-level likelihoods and priors based on unlabeled test samples.} These methods eliminate the need for backpropagation but remain constrained by CLIP's original feature space and pseudo-label quality.
However, most training-free TTA methods rely on dynamic cache mechanisms that prioritize high-confidence samples, assuming CLIP's features are sufficiently separated. This approach overlooks the potential of low-quality samples and is heavily dependent on pseudo-label quality, which can degrade performance when incorrect labels are cached. 
{To address this, MS-TTA enhances feature quality by leveraging all test samples in an unsupervised manner, thereby reducing the reliance on strict high-confidence selection and mitigating the influence of noisy pseudo-labels.}

\subsection{CLIP and Vision-Language Modeling}

Contrastive vision-language models like CLIP~\cite{clip} and ALIGN~\cite{ALIGN} enable zero-shot recognition and retrieval by aligning image and text embeddings. Their generalization across tasks has spurred developments in downstream applications \cite{dong2025beyond,downsteam1,downsteam2}. 
In camera-sensitive scenarios such as person re-identification, CLIP-based camera-agnostic feature learning can mitigate domain biases at training time~\cite{tan2025clip_cameraagnostic}. For pedestrian attribute recognition, both visual–textual baselines and transformer-based multi-task models have demonstrated robust representations under challenging conditions~\cite{cheng2022visualtext_pedestrian,fan2023parformer}. These lines of work are complementary to TTA: the former reduces domain shift via representation design during training, whereas our MS-TTA performs lightweight, parameter-free refinement at test time.

\subsection{Mean-shift and its Applications}
Mean-shift is a non-parametric technique for identifying the modes of a density function by iteratively shifting data points towards the weighted average of their neighbors~\cite{fukunaga1975estimation}. Its simplicity and effectiveness have made it widely applicable in clustering~\cite{meanshift,cheng1995mean}, object tracking~\cite{comaniciu2000real,kumar2022gridshift,jang2021meanshiftpp}, image segmentation~\cite{comaniciu2003kernel,kong2018recurrent}, and self-supervised learning~\cite{koohpayegani2021mean}. Other extensions include its application in theoretical analysis in mode-seeking behavior~\cite{cheng1995mean,fashing2005mean}, and advanced variants such as Von Mises-Fisher Mean Shift~\cite{kobayashi2010hypersphere}, GridShift~\cite{kumar2022gridshift}, and MeanShift++~\cite{jang2021meanshiftpp}. Additionally, Mean-shift has been studied in the context of mixture model modal clustering~\cite{chacon2019mixture}, convergence analysis~\cite{yamasaki2023convergence,li2007note}, and bound optimization~\cite{fashing2005mean}. Furthermore, recent advancements in Mean-shift have demonstrated its potential in robust probabilistic estimation~\cite{singh2004robust}, semi-supervised clustering~\cite{anand2013semi}, and agglomerative clustering~\cite{yuan2010agglomerative}, highlighting its adaptability to diverse problem settings. {These prior works have not explored Mean-Shift as a single-step, kNN-based refinement module for CLIP-based test-time adaptation. MS-TTA fills this gap by coupling such a refinement with a dynamic cache of refined embeddings in a fully training-free TTA framework.}
Despite its versatility, Mean-shift’s application in test-time adaptation (TTA) remains limited. Our work extends Mean-shift to improve feature alignment and clustering during test time, leveraging all available test samples, including low-confidence ones, to enhance performance. By integrating these insights, we propose a novel framework that combines the strengths of Mean-shift with modern machine-learning techniques to address the challenges of TTA.

\section{Preliminaries}
\label{sec:Preliminaries}

\subsection{Training Free Baseline}  
\label{sec: A Training Free Baseline}

CLIP~\cite{clip} is a pre-trained vision-language model composed of two parts: a visual encoder and a text encoder, which we represent separately \(E_{v}(\theta_v)\) and \(E_{t}(\theta_t)\). In classification tasks, given a test image \(x_{\text{test}}\) and \textit{N} classes, CLIP uses \(E_{t}(\theta_t)\) and \(E_{v}(\theta_v)\)  to encode handcrafted text descriptions of the \textit{N} classes and \(x_{\text{test}}\). After obtaining the corresponding text embeddings \( \mathbf{W}_t \) and {feature embeddings} \( \mathbf{f}_{\text{test}} \), CLIP matches the image with the most relevant text description to produce the final prediction as follows:
\begin{equation}
    \text{logits}_{\text{CLIP}} = \mathbf{f}_{\text{test}} \mathbf{W}_t^{\text{T}}.
    \label{eq:clip_logist}
\end{equation}

Before starting our method, we first construct a training-free baseline. We utilize a dynamic queue to store a set of representative samples and use these samples to assist in the prediction of test examples. This prediction is combined with the zero-shot CLIP predictions to produce the final inference. Specifically, we dynamically store \textbf{Q} test examples for each pseudo-class, along with their corresponding pseudo-labels \( \hat{y} \), using minimum entropy as the criterion. {For each pseudo-class we maintain a dynamic entropy threshold, defined as the highest entropy among the samples currently stored in its cache; a new test sample is inserted only if its prediction entropy is lower than this threshold, and if the cache is full it replaces the cached sample with the highest entropy.} The pseudo-labels are obtained by one-hot encoding the predictions \( \mathbf{f}_{\text{test}} \mathbf{W}_t^{\text{T}} \) for each sample:
\begin{equation}\label{eq:onehot}
    \hat{y} = \text{OneHot}(\mathbf{f}_{\text{test}} \mathbf{W}_t^{\text{T}}).
\end{equation}

When the queue reaches capacity \(Q\), we update the queue by replacing the sample with the highest entropy using the principle of minimizing entropy. This ensures that the cache always stores the most informative samples. Then, during testing, we retrieve the most relevant cache samples for each new test sample \( x_{\text{test}} \). For each unseen test sample, the \( E_v(\theta_v) \) generates the corresponding feature embedding \( \mathbf{f}_{\text{test}} \). The cache logits are then computed by retrieving the stored feature embeddings from the cache, and their relevance to the test sample is determined through a similarity measure, typically cosine similarity. The cache classifier aggregates the features of the stored cache samples, weighted by their similarity to the test sample’s features, to obtain the final cache logits. The prediction from the cache is computed as:
\begin{equation}
    \text{logits}_{\text{cache}}= \sum_{i=1}^{K} g(x_i)^\top g(x_{\text{test}}) \cdot \hat{y}_i,
\end{equation}
{where $g(x_i)$ represents the feature of each cached sample $x_i$, $K$ denotes the current number of cached items, and $\hat{y}_i$ denotes the test-time predicted label of the cached sample $x_i$.}

The final prediction is the combination of the cache classifier’s logits and the zero-shot CLIP logits:
\begin{equation}
    \text{logits}_{\text{final}} = \text{logits}_{\text{CLIP}} + \text{logits}_{\text{cache}}.
\end{equation}

By leveraging the cache and combining it with the zero-shot CLIP model’s predictions, our approach provides a training-free mechanism for test-time adaptation. This enables the model to adapt to unseen data and tasks dynamically during the test phase without retraining, making it effective in handling distribution shifts and unseen classes. {Compared with existing cache-based TTA methods such as TDA~\cite{tda} and BoostAdapter~\cite{Zhang2024BoostAdapter}, this training-free baseline adopts a simplified design: it maintains only a single positive cache that stores low-entropy (high-confidence) samples and uses cosine similarity to compute cache logits, rather than employing positive/negative caches. As such, it serves as a minimal internal reference within our framework, on top of which the proposed MS-TTA further introduces mean-shift refinement and dynamic cache updates.}

\begin{figure*}[t]
  \centering
   \includegraphics[width=0.99\linewidth]{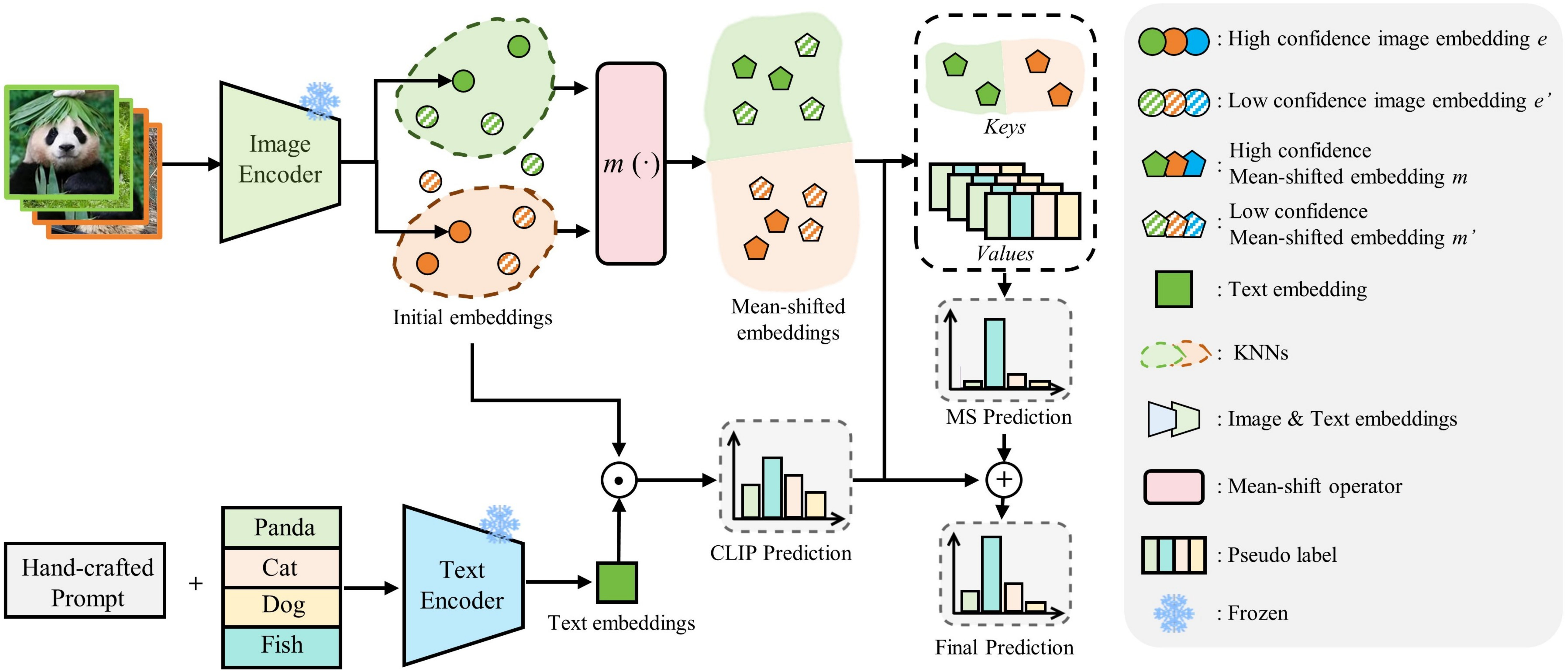}
   \caption{{\textbf{Overview of the MS-TTA}. We first extract initial embeddings using the CLIP {visual encoder} and refine them via a mean-shift operator with k-nearest neighbors ({kNN}), generating mean-shifted embeddings. These refined embeddings are dynamically stored in a key-value cache. During inference, CLIP predictions are combined with mean-shift-enhanced predictions, leveraging the cache to refine logits and improve classification accuracy, ensuring robustness to distribution shifts.}}
   \label{fig:NC-GCD}
\end{figure*}

\subsection{Mean-Shift Algorithm}  
The mean-shift algorithm is a non-parametric technique for locating the maxima of a density function in a feature space. Given a set of data points $\mathcal{V} = \{\mathbf{v}_1, \mathbf{v}_2, \dots, \mathbf{v}_n\}$, the algorithm shifts each point $\mathbf{v}$ toward the weighted mean of its neighborhood $\mathcal{N}(\mathbf{v}) \subseteq \mathcal{V}$. The weighted mean $m(\mathbf{v})$ is computed as:  
\begin{equation}  
    m(\mathbf{v}) = \frac{\sum_{\mathbf{v}_i \in \mathcal{N}(\mathbf{v})} \varphi(\|\mathbf{v}_i - \mathbf{v}\|) \mathbf{v}_i} {\sum_{\mathbf{v}_i \in \mathcal{N}(\mathbf{v})} \varphi(\|\mathbf{v}_i - \mathbf{v}\|)}, \label{eq:ms}  
\end{equation}  
where $\varphi(\cdot)$ is a kernel function that assigns weights based on the Euclidean distance $\|\mathbf{v}_i - \mathbf{v}\|$. The process iterates until convergence with the update rule:  
\begin{equation}  
    \mathbf{v}^{(t+1)} = m(\mathbf{v}^{(t)}), \label{eq:update}  
\end{equation}  
where $t$ denotes the iteration step.  
The algorithm's behavior depends on two key components. First, the neighborhood $\mathcal{N}(\mathbf{v})$ is defined by a fixed radius $h$, such that $\mathcal{N}(\mathbf{v}) = \{\mathbf{v}_i \in \mathcal{V} \mid \|\mathbf{v}_i - \mathbf{v}\| \leq h\}$. Second, the kernel function $\varphi(\cdot)$ assigns weights to neighboring points.

\section{Method}
\label{sec:method}
We integrate mean-shift clustering into test-time adaptation to refine feature embeddings beyond the original CLIP space. Using a single-step mean shift with k-nearest neighbors ({kNN}) (Sec.\ref{sec:mean_shifted_embedding}), we enhance feature consistency and robustness in a self-supervised manner. High-confidence mean-shifted embeddings are dynamically stored in a cache, which adapts by retaining low-entropy samples. During inference, the cache classifier retrieves relevant embeddings to compute cache-based logits (Sec.\ref{sec:Mean-Shifted Test-Time Adaptation}). The final prediction combines the zero-shot CLIP logits with the cache-enhanced logits, improving generalization to unseen distributions.

\subsection{Mean-Shifted Embedding}
\label{sec:mean_shifted_embedding}

Given a set of input images \(\mathcal{X} = \{x_1, x_2, \dots, x_N\}\), we extract their feature representations using a {visual encoder \(E_{v}\)}, producing a set of \( d \)-dimensional, \( l_2 \)-normalized embeddings:
\begin{equation}
{\mathcal{V} = \{ \mathbf{v}_i \}_{i=1}^{N}, \quad \text{where } \mathbf{v}_i = E_{v}(x_i).}
\end{equation}
To obtain high-quality feature representations in a self-supervised manner, we adopt a pre-trained CLIP {visual encoder}~\cite{clip}, though our method remains flexible and is not restricted to any particular backbone.

\subsubsection{ \textbf{Mean-Shifted Embedding Formulation}  }
Instead of directly using the raw embeddings, we refine them via a \textit{single-step mean shift} transformation, which adjusts each embedding towards the weighted mean of its local neighborhood. This process enhances feature discrimination and robustness. Unlike conventional mean shift, which selects neighbors based on a fixed-radius criterion, we employ \textit{k-nearest neighbors}, ensuring stable neighborhood selection and efficient GPU-based computation. The transformed embedding \(\mathbf{z}_i\) is defined as:
\begin{equation}
    \mathbf{z}_i = m(\mathbf{v}_i),
\end{equation}
where \( m(\cdot) \) denotes the mean shift operator.

\subsubsection{ \textbf{Neighborhood Definition}  }
For each embedding \(\mathbf{v}_i\), we define its local neighborhood \(\mathcal{N}(\mathbf{v}_i)\) as the set containing itself and its \( k \)-nearest neighbors based on cosine similarity:
\begin{equation}
    \mathcal{N}(\mathbf{v}_i) = \{\mathbf{v}_i\} \cup \operatorname{argmax}^k_{\mathbf{v}_j \in \mathcal{M}} \mathbf{v}_i \cdot \mathbf{v}_j,
    \label{eq:kNN}
\end{equation}
where \(\operatorname{argmax}^k\) retrieves the top-\( k \) neighbors that maximize the similarity measure and $\mathcal{M}$ stores feature embeddings of previously seen test samples.

\subsubsection{ \textbf{Kernel Weighting Strategy} } 
To control the contribution of each neighbor, we apply a kernel function \(\varphi(\cdot)\), which assigns higher importance to the central embedding \(\mathbf{v}_i\) while proportionally distributing weight among its neighbors:
\begin{align}
    \varphi(\mathbf{v}_j) &= 
    \begin{cases}
         1 - \alpha, & \text{if } \mathbf{v}_j = \mathbf{v}_i, \\
         \frac{\alpha}{k}, & \text{otherwise}.
    \end{cases}
    \label{eq:mean_shifted_embedding_meanvector}
\end{align}
Here, \(\alpha\) is a scaling factor that balances the influence of the original embedding and its neighbors. This formulation serves as an approximation of a simple uniform kernel weighting.

\subsubsection{ \textbf{Final Mean-Shifted Embedding Calculation}  }
The mean-shifted embedding \(\mathbf{z}_i\) is computed by aggregating the neighborhood embeddings according to their assigned weights, followed by \( l_2 \)-normalization to ensure unit norm:
\begin{equation}
    \mathbf{z}_i = \frac{\sum_{\mathbf{v}_j \in \mathcal{N}(\mathbf{v}_i)} \varphi(\mathbf{v}_j) \mathbf{v}_j}{\left\| \sum_{\mathbf{v}_j \in \mathcal{N}(\mathbf{v}_i)} \varphi(\mathbf{v}_j) \mathbf{v}_j \right\|}.
    \label{eq:mean_shifted_embedding}
\end{equation}
This step ensures that the refined embedding remains on the unit hypersphere while benefiting from local structural information.

These mean-shifted embeddings serve as enhanced feature representations, further improving downstream tasks such as classification. In subsequent sections, we explore how these embeddings integrate into Test-Time Adaptation and contribute to refining the overall adaptation process.

\begin{algorithm}[t!]
  \caption{Mean-Shifted Test-Time Adaptation}
  \label{alg:ms_tta}
  \begin{algorithmic}[1]
  \vspace{+1mm}
   \State \textbf{Input:} Test images $\mathcal{X} = \{x_1, x_2, \dots, x_N\}$, 
   CLIP visual encoder $E_v$, precomputed text embeddings $\mathbf{W}_t$, 
   number of nearest neighbors $k$, weighting factor $\alpha$, cache $\mathcal{C} = \emptyset$
   
   \State \textbf{Output:} Final predictions $\mathcal{P}$

   \For{each test sample $x_{\text{test}} \in \mathcal{X}$}
        \State Extract feature embedding:
        \[
        \mathbf{f}_{\text{test}} = E_v(x_{\text{test}})
        \] 
        \State Compute initial CLIP logits:
        \[
        \text{logits}_{\text{CLIP}} = \mathbf{f}_{\text{test}} \mathbf{W}_t^{\top}
        \]
        \State \textbf{Compute Mean-Shifted Embedding}
        \State Identify $k$-nearest neighbors:
        \[
        \mathcal{N}(\mathbf{f}_{\text{test}}) = \{ \mathbf{f}_{\text{test}} \} \cup \operatorname{argmax}^k_{\mathbf{f}_j \in \mathcal{M}} \mathbf{f}_{\text{test}} \cdot \mathbf{f}_j
        \]
        \State where $\mathcal{M}$ is the set of feature embeddings from previously seen test samples.
        \State Compute mean-shifted embedding:
        \[
        \mathbf{z}_{\text{test}} = \frac{(1-\alpha-\frac{\alpha}{k}) \mathbf{f}_{\text{test}} + \frac{\alpha}{k} \sum_{\mathbf{f}_j \in \mathcal{N}(\mathbf{f}_{\text{test}})} \mathbf{f}_j}
        {\left\| (1 - \alpha- \frac{\alpha}{k}) \mathbf{f}_{\text{test}} + \frac{\alpha}{k} \sum_{\mathbf{f}_j \in \mathcal{N}(\mathbf{f}_{\text{test}})} \mathbf{f}_j \right\|}.
        \]
        
        \State \textbf{Update Cache}
        \State Compute entropy $H(\text{logits}_{\text{CLIP}})$
        \State {$ threshold$ is the current highest entropy in the cache $\mathcal{C}_c$ for class $c$.}
        \If{$H(\text{logits}_{\text{CLIP}}) < \textit{threshold}$}
            \State Store $\mathbf{z}_{\text{test}}$ in cache $\mathcal{C}$
        \EndIf
        
        \State \textbf{Compute Mean-shift Logits}
        \If{$|\mathcal{C}| > 0$}
            \State Retrieve cache embeddings $\mathbf{Z}_{\text{cache}}$
            \State Compute similarity-based cache logits:
            \[
            \text{logits}_{\text{MS}} = \mathbf{z}_{\text{test}} \mathbf{Z}_{\text{cache}}^{\top} \mathbf{Y}
            \]
        \Else
            \State $\text{logits}_{\text{MS}} = \mathbf{0}$
        \EndIf
        
        \State \textbf{Final Prediction}
        \State Compute final logits:
        \[
        \text{logits}_{\text{final}} = \text{logits}_{\text{CLIP}} + \lambda \text{logits}_{\text{MS}}
        \]
        \State Obtain prediction:
        \[
        \hat{y} = \arg\max(\text{logits}_{\text{final}})
        \]
   \EndFor
   \State \Return Final predictions $\mathcal{P}$
   \vspace{+1mm}
   \end{algorithmic}
\end{algorithm}

\subsection{Mean-Shifted Test-Time Adaptation}
\label{sec:Mean-Shifted Test-Time Adaptation}

Given a test image \( x_{\text{test}} \), we first obtain its feature representation using the CLIP visual encoder \( E_v \), resulting in the test image embedding: $\mathbf{f}_{\text{test}} = E_v(x_{\text{test}})$.
The initial prediction logits $\text{\textit{logits}}_{\text{CLIP}}$ are computed by matching this embedding against the class-aligned text embeddings \( \mathbf{W}_t \) using Eq.\ref{eq:clip_logist}.

\subsubsection{ \textbf{Mean-Shifted Embedding Computation}  }
To refine the extracted test-time features, we apply a single-step mean shift operation, which enhances feature consistency by adjusting \( \mathbf{f}_{\text{test}} \) based on its k-nearest neighbors in the feature space.
The local neighborhood \( \mathcal{N}(\mathbf{f}_{\text{test}}) \) is defined as Eq.\ref{eq:kNN}.
According to Eq.\ref{eq:mean_shifted_embedding_meanvector} and Eq.\ref{eq:mean_shifted_embedding}, the mean-shifted embedding \( \mathbf{z}_{\text{test}} \) is then computed as:
{
\begin{equation}
    \mathbf{z}_{\text{test}} = \frac{(1-\alpha-\frac{\alpha}{k}) \mathbf{f}_{\text{test}} + \frac{\alpha}{k} \sum_{\mathbf{f}_j \in \mathcal{N}(\mathbf{f}_{\text{test}})} \mathbf{f}_j}
    {\left\| (1 - \alpha- \frac{\alpha}{k}) \mathbf{f}_{\text{test}} + \frac{\alpha}{k} \sum_{\mathbf{f}_j \in \mathcal{N}(\mathbf{f}_{\text{test}})} \mathbf{f}_j \right\|}.
\end{equation}
Here, the parameter \( \alpha \) balances the contribution of the test feature and its neighbors. The \(\alpha\) used here is the same hyperparameter as in Eq.~\ref{eq:mean_shifted_embedding_meanvector}.}

\subsubsection{ \textbf{Mean-shift Logits Computation}  }
If the entropy of the prediction is low, we store the mean-shifted embedding \( \mathbf{z}_{\text{test}} \) into a Mean-shift dynamic cache. The cache maintains a collection of previously observed embeddings, replacing the least confident entries based on entropy minimization.

For a new test sample, we retrieve stored mean-shifted embeddings from the cache and compute a similarity-based classification score. Given a cache consisting of embeddings \( \mathbf{Z}_{\text{cache}} \) and their corresponding pseudo-labels \( \mathbf{Y} \), we derive the Mean-shift enhanced logits as:
\begin{equation}
    \text{logits}_{\text{MS}} = \mathbf{z}_{\text{test}} \mathbf{Z}_{\text{cache}}^{\top} \mathbf{Y}.
\end{equation}
This step allows the model to incorporate prior knowledge stored in the cache to refine its predictions.
The final classification logits are obtained by linearly combining the original CLIP logits with the cache logits:
\begin{equation}
    \text{logits}_{\text{final}} = \text{logits}_{\text{CLIP}} + \lambda \text{logits}_{\text{MS}},
\end{equation}
where \( \lambda \) is a scaling factor that balances the contribution of the  Mean-shift enhanced prediction.

\subsection{Pseudocode for MS-TTA}
\label{sec:appendix_pseudocode}

To provide a clearer understanding of our approach, we present the pseudocode for MS-TTA in Algorithm~\ref{alg:ms_tta}. The pseudocode outlines the key steps of our adaptation framework, which refines test-time feature embeddings through mean-shift clustering and cache-based retrieval, enhancing model robustness under distribution shifts.

Unlike existing test-time adaptation methods that depend on pseudo-label quality or high-confidence sample selection, MS-TTA refines all test samples, including low-confidence ones, ensuring a self-supervised feature enhancement process. Additionally, as MS-TTA operates in a \textit{training-free} manner, it does not require parameter updates, making it computationally efficient while significantly improving adaptation performance.

\begin{table}[t]
\caption{Overview of datasets in the OOD benchmark and Cross-Dataset Benchmark, including the number of classes and test samples for each dataset.}
\centering
\small
\setlength{\tabcolsep}{5pt}
\begin{tabular}{llcc}
\toprule
\textbf{Benchmark} & \textbf{Dataset} & \textbf{Classes} & \textbf{Test Samples} \\
\midrule
\multirow{5}{*}{{OOD}}
&ImageNet & 1,000 & 50,000 \\
&ImageNet-V2 & 1,000 & 10,000 \\
&ImageNet-S & 1,000 & 50,000 \\
&ImageNet-A & 200 & 7,500 \\
&ImageNet-R & 200 & 30,000 \\
\midrule
\multirow{10}{*}{{Cross-Dataset}}
&Aircraft & 100 & 3,333 \\
&Caltech101 & 101 & 2,465 \\
&Cars & 196 & 8,041 \\
&DTD & 47 & 1,692 \\
&EuroSAT & 10 & 8,100 \\
&Flowers102 & 102 & 2,463 \\
&Food101 & 101 & 30,300 \\
&Pets & 37 & 3,669 \\
&SUN397 & 397 & 19,850 \\
&UCF101 & 101 & 3,783 \\
\bottomrule
\end{tabular}
\label{tab:cross_dataset}
\end{table}

\begin{table*}[t]
\caption{Full results on the {Cross-Dataset Benchmark} with ResNet50 and ViT-B/16 backbones. (a) shows results with ResNet50; (b) shows results with ViT-B/16. † indicates that this method is a training-free approach in test-time adaptation task.}
\centering
\small
\setlength{\tabcolsep}{3.2pt}
\begin{tabular}{l c c c c c c c c c c >{\columncolor{gray!20}}c}
\toprule
{Method} & \rotatebox{45}{Aircraft} & \rotatebox{45}{Caltech101} & \rotatebox{45}{EuroSAT} & \rotatebox{45}{Flowers102} & \rotatebox{45}{Oxford Pets} & \rotatebox{45}{SUN397} & \rotatebox{45}{UCF101} & \rotatebox{45}{Stanford Cars} & \rotatebox{45}{DTD} & \rotatebox{45}{Food101} & \rotatebox{45}{Average} \\
\midrule
\multicolumn{12}{l}{\textit{(a) Full results on the {Cross-Dataset Benchmark} with ResNet50 backbone}} \\
\midrule
CLIP          & 16.11 & 87.26 & 25.79 & 62.77 & 82.97 & 60.85 & 59.48 & 55.89 & 40.37 & 74.82 & 56.63 \\
CoOp          & 15.12 & 86.53 & 26.20 & 61.55 & \underline{87.00} & 58.15 & 59.05 & 55.32 & 37.29 & 75.59 & 56.18 \\
CoCoOp        & 14.61 & 86.38 & 28.73 & 65.57 & \textbf{88.39} & 59.61 & 57.10 & 56.22 & 38.53 & 76.20 & 57.13 \\
\cmidrule(r){1-12}
TPT (NeurIPS-22)  & 17.58 & 87.02 & 28.33 & 62.69 & 84.49 & 61.46 & 60.82 & 58.46 & 40.84 & 74.88 & 57.66 \\
DiffTPT (ICCV-23)       & 17.60 & 86.89 & 41.04 & 63.53 & 83.40 & 62.72 & 62.97 & \underline{60.71} & 40.72 & \underline{79.21} & 59.88 \\
HisTPT(NeurIPS-24) & 18.10 & 87.20 & 42.50 & 67.60 & 84.90 & \underline{63.50} & 64.10 & \textbf{61.30} & 41.30 & \textbf{81.30} & 61.18 \\
{DPE (NeurIPS-24)   }& {19.80}  &{\textbf{90.83}}  &{41.67}  & {67.60} & {85.97}& {\textbf{64.24}}  &{61.98}  &{59.26}  & {\textbf{50.18}} &{77.83}  & {\underline{61.93}} \\
\cmidrule(r){1-12}
TDA (CVPR-24)  †        & 17.61 & 89.70 & 42.11 & \underline{68.74} & 86.18 & 62.53 & 64.18 & 57.78 & 43.74 & 77.75 & 61.03 \\
BoostAdapter (NeurIPS-24) †&18.93 & 88.48 & \underline{44.40} & 68.25 & 85.75 & 62.83 & \underline{64.42} & 59.67 & 43.85 & 78.78 & 61.54 \\
BCA (CVPR-25) † & \textbf{19.89} & \underline{89.70} & 42.12 & 66.30 & 85.58 & 63.38 & 63.51 & 58.13 & \underline{48.58} & 77.19 & 61.44 \\
\rowcolor{gray!20} \textbf{MS-TTA (Ours) †} &  \underline{19.23} & 88.52 & \textbf{47.61} & \textbf{68.94} & 86.02 & 63.05 & \textbf{64.68} & 59.61 &  43.97 & 78.85 & \textbf{62.05} \\

\midrule
\multicolumn{12}{l}{\textit{(b) Full results on the {Cross-Dataset Benchmark} with ViT-B/16 backbone}} \\
\midrule
CLIP          & 23.22 & 93.55 & 50.42 & 66.99 & 86.92 & 65.63 & 65.16 & 66.11 & 45.04 & 82.86 & 64.59 \\
CoOp          & 18.47 & 93.70 & 46.39 & 68.71 & 89.14 & 64.15 & 66.55 & 64.51 & 41.92 & 85.30 & 63.88 \\
CoCoOp        & 22.29 & 93.79 & 39.23 & 70.85 & {90.46} & 66.89 & 68.44 & 64.90 & 45.45 & 83.97 & 64.63 \\
\cmidrule(r){1-12}
TPT (NeurIPS-22) & 24.78 & 94.16 & 42.44 & 68.98 & 87.79 & 65.50 & 68.04 & 66.87 & {47.75} & 84.67 & 65.10 \\
DiffTPT (ICCV-23) & 25.60 & 92.49 & 43.13 & 70.10 & 88.22 & 65.74 & 62.67 & 67.01 & 47.00 & {87.23} & 64.92 \\
MTA (CVPR-24) & 25.32 & 94.13 & 38.71 & 68.26 & 88.22 & 64.98 & 68.11 & 68.05 & 45.59 & 84.95 & 64.63 \\
MTA+Ensemble   & 25.20 & 94.21 & 45.36 & 68.06 & 88.24 & 66.67 & 68.69 & 68.47 & 45.90 & 85.00 & 65.58 \\
HisTPT (NeurIPS-24) & 26.90 & 94.50 & 49.70 & 71.20 & 89.10 & 67.20 & 70.10 & 69.20 & 48.90 & \textbf{89.30} & 67.61 \\
{DPE (NeurIPS-24)}& {\textbf{28.95}}  &{\underline{94.81}}  &{55.79}  & {\textbf{75.07}} & {91.14}& {\textbf{70.07}}  &{70.44}  &{67.31}  & {\textbf{54.20}} &{86.17}  & {\underline{69.40}} \\
\cmidrule(r){1-12}
TDA (CVPR-24) † & 23.91 & 94.24 & 58.00 & 71.42 & 88.63 & 67.62 & 70.66 & 67.28 & 47.40 & 86.14 & 67.53 \\
BCA (CVPR-25) † & \underline{28.59} & 94.69 & 56.63 & 73.12 & \textbf{90.43} & \underline{68.41} & 67.59 & 66.86 &\underline{53.49} & 85.97 & 68.59 \\
BoostAdapter (NeurIPS-24) † & 27.45 & 94.77 & \underline{61.22} & {71.66} & {89.51} & {68.09} & \underline{71.93} & \underline{69.30} & 45.69 & 87.17 & 68.68 \\
\rowcolor{gray!20} \textbf{MS-TTA (Ours) †} & 27.78 & \textbf{95.01} & \textbf{65.21} & \underline{73.20} & \underline{90.11} & \textbf{68.42} & \textbf{72.38} & {\textbf{69.49}} & 45.86 & \underline{87.38} & \textbf{69.48}\\

\rowcolor{gray!40} \textit{\textbf{Improv over BoostAdapter}} & \textbf{+0.33} & \textbf{+0.24} & \textbf{+3.99} & \textbf{+1.54} & \textbf{+0.60} & \textbf{+0.33} & \textbf{+0.45} & \textbf{{+0.19}} & \textbf{+0.17} & \textbf{+0.21} & \textbf{+0.80} \\
\bottomrule
\end{tabular}
\label{tab:full_results}
\end{table*}

\section{Experiment} \label{sec:Experiment}

\subsection{Experimental Setup} \label{sec:setup}

\subsubsection{\textbf{Benchmarks}}
We evaluate our method using two key benchmarks: the out-of-distribution (OOD) benchmark and the Cross-Dataset Benchmark, same as prior work~\cite{tpt,difftpt,tda}.

\noindent \textbf{OOD Benchmark.}
We evaluate the model’s robustness to distributional shifts using the Out-of-Distribution (OOD) benchmark, which includes ImageNet~\cite{imagenet} and four challenging variants: ImageNet-A~\cite{imageneta}, ImageNet-R~\cite{imagenetr}, ImageNet-V2~\cite{imagenetv2}, and ImageNet-S~\cite{imagenetsk}.

\noindent \textbf{Cross-Dataset Benchmark.}  
To further assess generalization across domains, we employ a Cross-Dataset Benchmark spanning 10 datasets from diverse visual categories. These include general object classification (Caltech101~\cite{caltech101}), fine-grained recognition (OxfordPets~\cite{pets}, StanfordCars~\cite{cars}, Flowers102~\cite{flowers}, Food101~\cite{food101}, Aircraft~\cite{aircraft}), scene and texture understanding (SUN397~\cite{sun397}, DTD~\cite{dtd}), and specialized domains such as satellite imagery (EuroSAT~\cite{eurosat}) and video-based action recognition (UCF101~\cite{ucf101}). This benchmark evaluates the model’s ability to transfer knowledge to entirely different domains with distinct semantics, styles, and modalities. Dataset statistics are provided in Table~\ref{tab:cross_dataset}.

\subsubsection{\textbf{Comparison Methods}}
{We compare our approach with several SOTA methods in the test-time adaptation (TTA) domain, including zero-shot CLIP~\cite{clip}, CoOp~\cite{coop}, Tip-Adapter~\cite{tip}, CoCoOp~\cite{cocoop}, TPT~\cite{tpt}, DiffTPT~\cite{difftpt}, HisTPT~\cite{Zhang2024Historical}, MTA~\cite{mta}, DPE~\cite{DPE}, as well as training-free TTA methods such as TDA~\cite{tda}, BCA~\cite{BCA} and BoostAdapter~\cite{Zhang2024BoostAdapter}.} However, Tip-Adapter is excluded from the Cross-Dataset Benchmark due to its inability to handle unseen classes during testing. Additionally, we do not compare with MTA in experiments using the ResNet50 backbone, as there is no data available for MTA on this architecture. The ensemble prediction method from MTA is referred to as MTA+Ensemble. Importantly, while TPT, DiffTPT, MTA, TDA and BoostAdapter operate within the original CLIP feature space, our method extends beyond this feature space.

\subsubsection{\textbf{Implementation Details}}
Our method is built on pre-trained CLIP~\cite{clip}, where the text encoder is a Transformer~\cite{attention_is_all_you_need} and the {visual encoder} is either ResNet50~\cite{resnet} or ViT-B/16~\cite{vit}. We use publicly available CLIP checkpoints, and all text prompts are manually defined and kept fixed. For MS-TTA, the main hyperparameters are the number of neighbors \(k\), the mean-shift weight \(\alpha\), {the logit scaling factor \(\lambda\)} and the cache capacity \(Q\); we choose \(k\) and \(\alpha\) based on the ablation studies, and set the cache capacity \(Q\) and the logit scaling factor \(\lambda\) to be consistent with the settings in TDA~\cite{tda}. We perform test-time adaptation with a batch size of 1 and conduct all experiments on a single NVIDIA RTX 3090 GPU.

\subsection{Comparison with State-of-the-Art Methods} \label{sec:compare}

We compare our approach with several State-of-the-Art methods, including zero-shot CLIP, CoOp, CoCoOp, Tip-Adapter, TPT, DiffTPT, HisTPT, MTA, DPE, BCA, BoostAdapter and TDA. It is important to note that Tip-Adapter cannot handle unseen classes during testing, limiting its evaluation on the Cross-Dataset Benchmark. Additionally, MTA does not provide accuracy results for experiments using the ResNet50 backbone. Like TPT, DiffTPT, MTA, and TDA, we evaluate our method on both the \textbf{OOD benchmark} and the \textbf{Cross-Dataset Benchmark} to assess its performance across diverse tasks and datasets.

\subsubsection{\textbf{Results on the {Cross-Dataset Benchmark}}}
Our method, MS-TTA, demonstrates impressive results on the {Cross-Dataset Benchmark}, significantly outperforming existing training-free test-time adaptation methods. As shown in Table \ref{tab:full_results}, MS-TTA consistently leads across multiple datasets, showing superior robustness to distribution shifts and better adaptation capabilities without the need for training.

As shown in Table \ref{tab:full_results}b, when using the ViT-B/16 backbone, MS-TTA achieves remarkable results, surpassing all training-free methods on 7 out of 10 datasets. Notably, MS-TTA shows an average accuracy improvement of \textbf{+0.80\%} over BoostAdapter. In particular, on datasets such as EuroSAT, MS-TTA improves by \textbf{+3.99\%} over BoostAdapter, highlighting its effectiveness in handling challenging domains. Additionally, it outperforms BoostAdapter on UCF101 and SUN397, demonstrating its versatility across a wide range of datasets, further proving its capability to generalize across different domains.

As shown in Table \ref{tab:full_results}a, on the ResNet50 backbone, MS-TTA achieves the highest average accuracy among all existing methods. Specifically, MS-TTA achieves leading results on datasets like UCF101, with substantial improvements over BoostAdapter and other methods. This shows that MS-TTA is not only effective with ViT-B/16 but also performs excellently with the ResNet50 backbone, further validating its versatility.

\subsubsection{\textbf{Results on the Out-of-Distribution Benchmark}}
Table~\ref{tab:method_comparison_ood} presents the performance of MS-TTA on the OOD benchmark using the ViT-B/16 backbone, while Table~\ref{tab:method_comparison_cross} shows the results with the ResNet50 backbone.

In both cases, MS-TTA achieves better or comparable performance than existing training-free methods across all OOD datasets. On the ViT-B/16 backbone, our method demonstrates superior performance to other training-free methods on each individual dataset, with a higher average accuracy compared to all training-free methods. Similarly, with the ResNet50 backbone, MS-TTA leads in performance on most datasets among training-free methods, further highlighting its robustness. The average accuracy also surpasses the competing methods in both backbones, validating the effectiveness of MS-TTA in adapting to unseen data distributions. These results reinforce the strong capabilities of MS-TTA in addressing distribution shifts and ensuring stable performance across various benchmarks.

\begin{table}[t]
\caption{Performance comparison across different methods with ViT-B/16 backbone. † indicates that this method is a training-free approach in test-time adaptation task.}
\centering
\small
\setlength{\tabcolsep}{4pt}
\begin{tabular}{l c c c c >{\columncolor{gray!20}}c}
\toprule
Method & {A} & {R} & {S} & {V2} & {Avg} \\
\midrule
CLIP & 49.89 & 77.65 & 48.24 & 61.88 & 59.42 \\
CoOp & 49.71 & 75.21 & 47.99 & 64.20 & 59.28 \\
CoCoOp & 50.63 & 76.18 & 48.75 & 64.07 & 59.91 \\
Tip-Adapter & 51.04 & 77.76 & 48.88 & 63.41 & 60.27 \\
\midrule
TPT (NeurIPS-22)  & 54.77 & 77.06 & 47.94 & 63.45 & 60.81 \\
DiffTPT (ICCV-23)  & 55.68 & 75.00 & 46.80 & 65.10 & 60.65 \\
MTA (CVPR-24)  & 57.41 & 76.92 & 48.58 & 63.61 & 61.63 \\
MTA+Ensemble  & 58.06 & 78.33 & 49.61 & 64.24 & 62.56 \\
{DPE (NeurIPS-24)}& {59.63}  &{\underline{80.40}}  &{\textbf{52.26}}  & {65.44} & {64.43} \\
\midrule
TDA (CVPR-24) †& 60.11 & 80.24 & 50.54 & 64.67 & 63.89 \\
BCA (CVPR-25) †& 61.14 & 80.72 & 50.87 & 64.90 & 64.41 \\
BoostAdapter (NeurIPS-24)   †& \underline{64.53} & 80.95 & 51.28 & \underline{65.51} & \underline{65.57} \\
\rowcolor{gray!20} \textbf{MS-TTA (Ours) †} & \textbf{64.63} & \textbf{81.08} & \underline{51.55} & \textbf{65.57} & \textbf{65.71} \\
\bottomrule
\end{tabular}
\label{tab:method_comparison_ood}
\end{table}

\begin{table}[t]
\caption{Performance comparison across different methods with ResNet50 backbones. † indicates that this method is a training-free approach in test-time adaptation task.}
\centering
\small
\setlength{\tabcolsep}{4pt}
\begin{tabular}{l c c c c >{\columncolor{gray!20}}c}
\toprule
Method & {A} & {R} & {S} & {V2} & {Avg} \\
\midrule
CLIP & 23.24 & 60.72 & 35.48 & 52.91 & 43.09 \\
CoOp & 23.06 & 56.60 & 34.67 & 55.40 & 42.43 \\
CoCoOp & 23.32 & 57.74 & 34.48 & 55.72 & 42.82 \\
Tip-Adapter & 23.13 & 60.35 & 35.74 & 53.97 & 43.30 \\
\midrule
TPT (NeurIPS-22)& 26.67 & 59.11 & 35.09 & 54.70 & 43.89 \\
DiffTPT (ICCV-23)& 31.06 & 58.80 & 37.10 & 55.80 & 45.69 \\
{DPE (NeurIPS-24)}& {30.15}  &{\textbf{63.72}}  &{\textbf{40.03}}  & {\textbf{56.72}} & {47.66} \\
\midrule
TDA (CVPR-24)†& 30.29 & 62.58 & 38.12 & 55.54 & 46.63 \\
BCA (CVPR-25)†& 30.35 & 62.89 & 38.08 & 56.58 & 46.98 \\
BoostAdapter (NeurIPS-24)† & \underline{35.12} & 62.66 & 38.87 & 56.14 & \underline{48.20} \\
\rowcolor{gray!20} \textbf{MS-TTA (Ours) †}& \textbf{35.62} & \underline{62.84} & \underline{39.10} & \underline{56.58} & \textbf{48.54} \\
\bottomrule
\end{tabular}
\label{tab:method_comparison_cross}
\end{table}

\begin{table}[t]
\caption{Performance comparison on Flowers102 and ImageNet-A across different values of K. The best values are in bold, and the second-best values are underlined.}
\centering
\small
\setlength{\tabcolsep}{12pt} 
\begin{tabular}{c c c}
\toprule
kNN numbers $k$ & Flowers102 & ImageNet-A \\
\midrule
\rowcolor{gray!20} 2  & \textbf{72.92}  & \textbf{64.63} \\
4  & \underline{72.47}  & 64.38 \\
6  & 72.38  & \underline{64.41} \\
8  & 72.35  & 64.47 \\
16 & 71.90  & 64.45 \\
\bottomrule
\end{tabular}
\label{tab:k_vs_performance}
\end{table}

\begin{table}[t]
\caption{Performance comparison with different MS Weights \( \alpha \) on DTD, ImageNet-A, and Flowers102 datasets. The best values are in bold, and the second-best values are underlined.}
\centering
\small
\setlength{\tabcolsep}{10pt} 
\begin{tabular}{c c c c}
\toprule
MS Weight \( \alpha \)& DTD & ImageNet-A & Flowers102 \\
\midrule
0   & 45.30  & 64.36  & 72.23  \\
0.2 & 45.32  & 64.44  & 72.43  \\
0.3 & 45.32  & 64.47  & 72.40  \\
0.4 & 45.44  & 64.45  & 72.39  \\
0.5 & 45.21  & 64.50  & 72.51  \\
0.6 & 45.26  & 64.55  & 72.47  \\
0.7 & 45.68  & \textbf{64.63}  & 72.72  \\
\rowcolor{gray!30} 0.8 & \textbf{45.86}  & \underline{64.60}  & \textbf{72.92}  \\
0.9 & \underline{45.74}  & 64.59  & \underline{72.89}  \\
1.0   & 45.58  & 64.57  & 72.83  \\
\bottomrule
\end{tabular}
\label{tab:ms_weight_comparison}
\end{table}

\begin{table*}[t]
\caption{Comparison of baseline and MS-TTA across different datasets.}
\centering
\small
\setlength{\tabcolsep}{7pt}
\begin{tabular}{l c c c c c c c c c c >{\columncolor{gray!20}}c}
\toprule
Method & \rotatebox{45}{Aircraft} & \rotatebox{45}{Caltech101} & \rotatebox{45}{Cars} & \rotatebox{45}{DTD} & \rotatebox{45}{EuroSAT} & \rotatebox{45}{Flowers} & \rotatebox{45}{Food101} & \rotatebox{45}{Pets} & \rotatebox{45}{SUN397} & \rotatebox{45}{UCF101} & \rotatebox{45}{Average} \\
\midrule
Baseline & 27.18 & 94.58 & 69.15 & 45.30 & 61.02 & 72.23 & 87.02 & 89.36 & 67.94 & 71.78 & 68.56 \\ \textbf{Ours(MS-TTA)} & \textbf{27.78} & \textbf{95.01} & \textbf{69.49} & \textbf{45.86} & \textbf{65.21} & \textbf{72.92} & \textbf{87.38} & \textbf{90.11} & \textbf{68.42} & \textbf{72.38} & \textbf{69.46 }\\
\midrule
\rowcolor{gray!30} \textbf{\textit{Improvement}} & \textbf{+0.60} & \textbf{+0.43} & \textbf{+0.34} & \textbf{+0.56} & \textbf{+4.19} & \textbf{+0.69} & \textbf{+0.36} & \textbf{+0.75} & \textbf{+0.48} & \textbf{+0.60} & \textbf{+0.90} \\
\bottomrule
\end{tabular}
\label{tab:improvement}
\end{table*}

\subsection{Ablation Studies}
In this section, we conduct ablation experiments to analyze the effectiveness of our design. Our baseline method is the one mentioned in Section ~\ref{sec: A Training Free Baseline}.

\begin{table}[t]
\caption{Comparison of baseline and MS-TTA on OOD benchmark.}
\centering
\small
\setlength{\tabcolsep}{4.8pt}
\begin{tabular}{l c c c c >{\columncolor{gray!20}}c}
\toprule
Method & {A} & {R} & {S} &{V2} & {OOD Avg} \\
\midrule
baseline & 64.36 & 80.11 & 49.89 & 65.11 & 64.87 \\
\textbf{Ours(MS-TTA)} & \textbf{64.63} & \textbf{81.08} & \textbf{51.55} & \textbf{65.57} & \textbf{65.71} \\
\midrule
\rowcolor{gray!30} \textbf{\textit{Improvement}} & \textbf{+0.27} & \textbf{+0.97} & \textbf{+1.66} & \textbf{+0.46} & \textbf{+0.84} \\
\bottomrule
\end{tabular}
\label{tab:improvement_boost}
\end{table}

\begin{table}[t]
\centering
\small
\caption{{Results on Flowers102, EuroSAT, Oxford Pets and ImageNet-based OOD benchmarks with 5 random seeds.}}
\setlength{\tabcolsep}{2.5pt}
\begin{tabular}{lcc >{\columncolor{gray!20}}c}
\toprule
{Dataset} & {BoostAdapter} & {MS-TTA (Mean $\pm$ Std)} & {Improv.} \\
\midrule
{EuroSAT}    & {61.22} & {65.08 $\pm$ 0.18} & {\textbf{+3.86}} \\
{Flowers}    & {71.66} & {73.21 $\pm$ 0.22} & {\textbf{+1.55}} \\
{Ox-Pets}    & {89.51} & {90.08 $\pm$ 0.08} & {\textbf{+0.57}} \\
\midrule
{ImageNet-A}  & {35.12} & {35.63 $\pm$ 0.09} & {\textbf{+0.51}} \\
{ImageNet-S}  & {38.87} & {39.12 $\pm$ 0.06} & {\textbf{+0.25}} \\
{ImageNet-V2} & {56.14} & {56.60 $\pm$ 0.12} & {\textbf{+0.46}} \\
\bottomrule
\end{tabular}
\label{tab:stat_results}
\end{table}

\begin{table}[t]
\centering
\small
\caption{{Effect of the number of Mean-Shift steps on accuracy (\%) and inference speed (FPS). ``Baseline'' denotes our cache baseline without Mean-Shift. Inference speed is measured on a single NVIDIA RTX 3090 GPU.}}
\setlength{\tabcolsep}{3pt}
\begin{tabular}{lcccc}
\toprule
{Metric} 
& {Baseline} 
& {Single-step} 
& {Two-steps} 
& {Four-steps} \\
\midrule
{ImageNet-R Acc.} 
& {80.11} 
& {\textbf{81.08}} 
& {\underline{80.80}} 
& {80.25} \\
{ImageNet-S Acc.} 
& {49.89} 
& {\textbf{51.55}} 
& {\underline{50.88}} 
& {49.95} \\
{EuroSAT Acc.} 
& {61.02} 
& {\textbf{65.21}} 
& {\underline{63.57}} 
& {60.92} \\
{Flowers Acc.} 
& {72.23} 
& {\textbf{72.92}} 
& {72.30} 
& {\underline{72.53}} \\
{Average Acc.} 
& {65.81} 
& {\textbf{67.69}} 
& {\underline{66.89}} 
& {65.91} \\
{Inference Speed} 
& {\textbf{12.33}} 
& {\underline{10.05}} 
& {8.47} 
& {5.13} \\
\bottomrule
\end{tabular}
\label{tab:mean_shift_steps}
\end{table}

\begin{table}[t]
\centering
\small
\caption{{Sensitivity analysis of the logit scaling factor $\lambda$ on Flowers102 and ImageNet-A.}}
\setlength{\tabcolsep}{12pt}
\begin{tabular}{c c c}
\toprule
{Scaling Factor $\lambda$} & {Flowers102} & {ImageNet-A} \\
\midrule
{0}   & {66.99} & {49.86} \\
{0.5} & {72.11} & {64.30} \\
{1}   & {\textbf{72.92}} & {\textbf{64.63}} \\
{2}   & {72.63} & {63.96} \\
{10}  & {69.18} & {63.77} \\
\bottomrule
\end{tabular}
\label{tab:lambda_vs_performance}
\end{table}

\subsubsection{\textbf{Effectiveness of MS-TTA}}
We first evaluate the effectiveness of our proposed MS-TTA by comparing it with the baseline method. Table~\ref{tab:improvement} presents the results across multiple datasets, showing a consistent improvement in accuracy with MS-TTA. On the 10 datasets, MS-TTA outperforms the baseline by an average of \textbf{0.90\%}, with significant gains in datasets such as EuroSAT (\textbf{+4.19\%}) and Pets (\textbf{+0.75\%}).
In the OOD benchmark (Table~\ref{tab:improvement_boost}), MS-TTA also demonstrates an advantage over the baseline, achieving an overall improvement of \textbf{+0.84\%}. The improvement is especially notable in tasks involving higher distribution shifts, such as in the "ImageNet-S", where MS-TTA boosts accuracy by \textbf{+1.66\%}.
These results highlight the effectiveness of MS-TTA in enhancing feature quality and robustness, and demonstrate that incorporating mean-shift clustering consistently improves the model’s performance without the need for retraining.

\subsubsection{\textbf{Impact of kNN numbers $k$}}
The number of nearest neighbors \( k \) plays a crucial role in our mean-shift-based test-time adaptation framework, directly influencing feature refinement and adaptation effectiveness. To analyze its impact, we conduct an ablation study across different \( k \) values on Flowers102 and ImageNet-A datasets, as shown in Table~\ref{tab:k_vs_performance}. The results indicate that setting \( k = 2 \) achieves the highest performance on both datasets, with an accuracy of \textbf{72.92\%} on Flowers102 and \textbf{64.63\%} on ImageNet-A. While increasing \( k \) to 4 or 6 maintains competitive results, larger values such as 8 or 16 lead to performance degradation. This suggests that while a larger \( k \) may incorporate more contextual information, it also introduces noise, weakening the effectiveness of feature refinement. The choice of \( k \) affects the balance between local compactness and global adaptability. A smaller \( k \) encourages stronger local feature consistency, allowing embeddings to shift toward more compact clusters and enhancing class separability. However, setting \( k \) too low may restrict adaptation in complex distributions, where a broader neighborhood is beneficial for capturing a more representative structure. Conversely, larger \( k \) values increase the inclusion of diverse features but risk diluting distinctive characteristics, making the mean-shift process less effective in refining features.

\subsubsection{\textbf{Effectiveness of MS scaling factor \( \alpha \)} } 

The mean-shift scaling factor \( \alpha \) plays a critical role in controlling the influence of the original feature embedding and its nearest neighbors in our test-time adaptation framework. A lower \( \alpha \) retains more of the original feature representation, while a higher \( \alpha \) increases the contribution of neighboring embeddings, thereby enhancing feature refinement. To investigate the impact of \( \alpha \), we conduct experiments on DTD, ImageNet-A, and Flowers102 datasets, with results summarized in Table~\ref{tab:ms_weight_comparison}.

From the table, we observe that performance improves as \( \alpha \) increases, reaching an optimal range around 0.7 to 0.9, before slightly declining at \( \alpha = 1.0 \). Specifically, the best accuracy is achieved at \( \alpha = 0.8 \) for DTD and Flowers102, while ImageNet-A performs best at \( \alpha = 0.7 \). This trend suggests that incorporating more contextual information from the local neighborhood helps refine embeddings, but excessive reliance on neighbor aggregation may introduce noise, leading to minor performance degradation at higher \( \alpha \) values.
The results reinforce our findings that moderate values of \( \alpha \) effectively balance original feature retention and local feature refinement, leading to improved adaptation performance. These insights suggest that selecting an appropriate \( \alpha \) is crucial for optimizing mean-shift-based test-time adaptation and could potentially be further improved through adaptive scaling strategies tailored to dataset characteristics.

\begin{table*}[t]
\caption{Full results on the {Cross-Dataset Benchmark} integrated with BoostAdapter and TDA. (a) shows results integrated with TDA; (b) shows results integrated with BoostAdapter. † indicates that this method is a training-free approach.}
\centering
\small
\setlength{\tabcolsep}{3pt}
\begin{tabular}{l c c c c c c c c c c >{\columncolor{gray!20}}c}
\toprule
{Method} & \rotatebox{45}{Aircraft} & \rotatebox{45}{Caltech101} & \rotatebox{45}{EuroSAT} & \rotatebox{45}{Flowers102} & \rotatebox{45}{Oxford Pets} & \rotatebox{45}{SUN397} & \rotatebox{45}{UCF101} & \rotatebox{45}{Stanford Cars} & \rotatebox{45}{DTD} & \rotatebox{45}{Food101} & \rotatebox{45}{Average} \\
\midrule
\multicolumn{12}{l}{\textit{ Full results on the {Cross-Dataset Benchmark} with ViT-B/16 backbone integrated with TDA}} \\
\cmidrule(r){1-12}
TDA (CVPR-24) † & 23.91 & 94.24 & 58.00 & 71.42 & 88.63 & 67.62 & 70.66 & 67.28 & 47.40 & 86.14 & 67.53 \\
\rowcolor{gray!20} \textbf{TDA + MS †} & \textbf{25.14} & \textbf{94.29} & \textbf{61.93} & \textbf{74.06} & \textbf{{90.08}} & \textbf{67.83} & \textbf{71.77} & {\textbf{67.33}} & \textbf{48.93} & \textbf{{86.35}} & \textbf{68.77} \\
\rowcolor{gray!30} \textit{\textbf{Improv over TDA}} & \textbf{+1.23} & \textbf{+0.05} & \textbf{+3.93} & \textbf{+2.64} & \textbf{+1.45} & \textbf{+0.21} & \textbf{+1.11} & \textbf{{+0.05}} & \textbf{+1.53} & \textbf{+0.21} & \textbf{+1.24} \\
\midrule
\multicolumn{12}{l}{\textit{ Full results on the {Cross-Dataset Benchmark} with ViT-B/16 backbone integrated with BoostAdapter}} \\
\midrule
BoostAdapter (NeurIPS-24) † & {27.45} & {94.77} & {61.22} & {71.66} & {89.51} & {68.09} & {71.93} & {69.30} & 45.69 & 87.17 &{68.68} \\
\rowcolor{gray!20} \textbf{BoostAdapter + MS †} & \textbf{27.76} & \textbf{95.03} & \textbf{64.82} & \textbf{72.73} & \textbf{{90.17}} & \textbf{68.53} & \textbf{72.44} & {\textbf{69.54}} & \textbf{45.91} &\textbf{87.41} & \textbf{69.43} \\
\rowcolor{gray!30} \textit{\textbf{Improv over BoostAdapter}} 
& \textbf{+0.31} & \textbf{+0.26} & \textbf{+3.60} & \textbf{+1.07} 
& \textbf{+0.66} & \textbf{+0.44} & \textbf{+0.51} & \textbf{+0.24} 
& \textbf{+0.22} & \textbf{+0.24} & \textbf{+0.75} \\
\bottomrule
\end{tabular}
\label{tab:full_results_pp}
\end{table*}

\begin{figure}[t]
  \centering
    \includegraphics[width=0.98\linewidth]{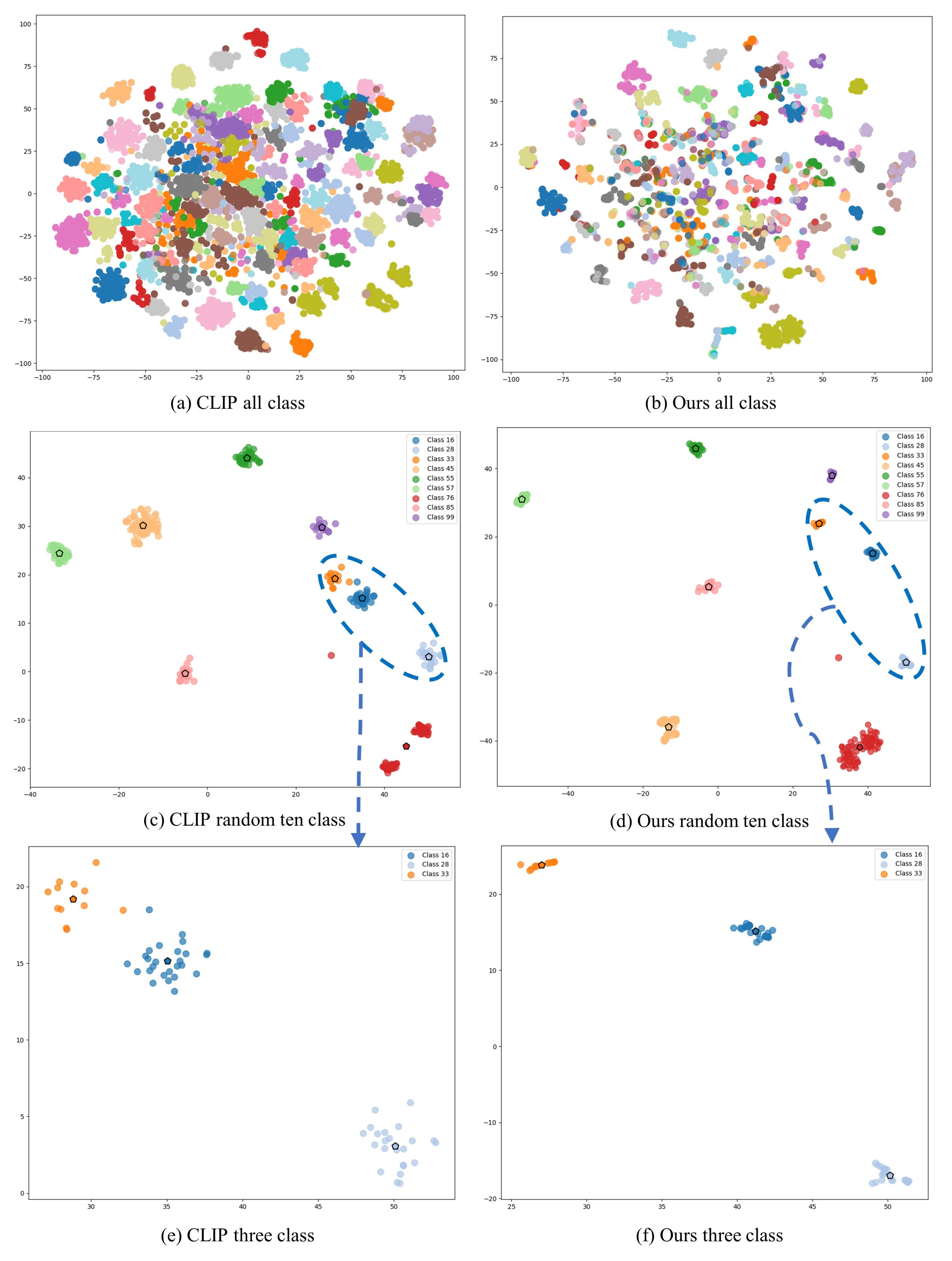}
  \caption{T-SNE visualizations of feature embeddings on the Flowers102 dataset. (a)-(b): Comparison of global embedding distributions from CLIP (a) and our method (b). (c)-(d): A focused view on 10 randomly selected classes, showing that our mean-shifted embeddings (d) reduce intra-class variance and enlarge inter-class margins compared to CLIP (c). (e)-(f): A close-up view of class 16 and class 33, where our method (f) achieves clearer separation and sharper decision boundaries than CLIP (e).}
  \label{fig:visualization}
\end{figure}

\subsubsection{{\textbf{Statistical Robustness over Random Seeds}}}

{To assess the sensitivity of MS-TTA to the online inference order, we further conduct experiments with 5 random seeds on six datasets, including three Cross-Dataset Benchmarks (EuroSAT, Flowers102, Oxford Pets) with ViT-B/16 backbone and three ImageNet-based OOD benchmarks (ImageNet-A, ImageNet-S, ImageNet-V2) with ResNet50 backbone. For each seed, we randomly shuffle the test samples within each dataset so that, under batch size 1, the evolving cache observes a different sequence of inputs during the TTA process.}

{The aggregated results are summarized in Table~\ref{tab:stat_results}, where we report the mean and standard deviation of the accuracy over 5 runs for MS-TTA, together with the single-run performance of BoostAdapter. MS-TTA consistently outperforms BoostAdapter on all six datasets, with improvements ranging from \textbf{+0.25\%} to \textbf{+3.86\%}, while the standard deviations remain at most 0.22. These observations confirm that the performance gains of MS-TTA are statistically significant and that our method is robust to variations in the online inference order, even in the early stage when the cache is still being populated.}

\subsubsection{{\textbf{Effect of Mean-shift Steps}}}
{To analyze the impact of applying Mean-shift multiple steps, we compare the baseline without Mean-shift with MS-TTA variants using 1, 2, and 4 Mean-shift steps on four representative datasets: ImageNet-R, ImageNet-S, EuroSAT, and Flowers102. For each configuration, we report top-1 accuracy, the average accuracy (Average Acc.), and inference throughput (FPS) in Table~\ref{tab:mean_shift_steps}. The single-step variant already brings a clear accuracy gain over the baseline, while the two-step setting remains slightly better than the baseline but worse than the single-step one, and the four-step variant brings almost no additional benefit while further reducing throughput. We therefore adopt the single-step kNN Mean-shift in MS-TTA as a good balance between effectiveness and efficiency in the training-free TTA setting.}

\begin{table}[t]
\caption{Comparison of inference speed, memory consumption, OOD results, and Cross-Dataset results across different methods on a NVIDIA 3090 GPU.}
\centering
\small
\setlength{\tabcolsep}{2pt}
\begin{tabular}{l c c c c}
\toprule
Method & \shortstack{Inference Speed \\(fps)} & \shortstack{Memory \\ (GB) }& \shortstack{OOD \\ Results} & \shortstack{Cross-Dataset \\Results} \\
\midrule
CLIP & \textbf{82.3} & \textbf{0.7} & 59.42 & 64.59 \\
TPT & 0.29 & 4.5 & 60.81 & 65.10 \\
DiffTPT & 0.10 & 14.4 & 60.65 & 64.92 \\
TDA & 11.89 & 1.2 & 63.89 & 67.53 \\
BoostAdapter & 11.23 & 1.2 & 65.57 & 68.68 \\
\rowcolor{gray!20} Ours & 10.05 & 1.4 & \textbf{65.71} & \textbf{69.48} \\
\bottomrule
\end{tabular}
\label{tab:performance_comparison}
\end{table}

\subsubsection{{\textbf{{Sensitivity of the logit scaling factor $\lambda$}}}}
{We analyze the sensitivity of the logit scaling factor $\lambda$, which controls the trade-off between the CLIP logits and the cache-based logits. In our implementation, we keep $\lambda$ consistent with the setting in TDA~\cite{tda} and sweep $\lambda \in \{0, 0.5, 1, 2, 10\}$ on Flowers102 and ImageNet-A. Table~\ref{tab:lambda_vs_performance} shows that MS-TTA yields consistently strong performance for moderate $\lambda$ values, with the best results achieved at $\lambda=1$ on both datasets. In contrast, $\lambda=0$ disables the cache contribution and leads to a clear drop, while a large value $\lambda=10$ over-emphasizes cache logits. Overall, the results indicate that MS-TTA is robust to $\lambda$ and does not require careful tuning of this parameter.}

\subsection{Visualization}
We use t-SNE visualization to illustrate the effectiveness of our proposed method, especially in enhancing feature discriminability. As shown in Fig.\ref{fig:visualization}, we compare the feature embeddings generated by CLIP and our method across different scenarios using the Flowers102 dataset. In Fig.\ref{fig:visualization}(a), the embeddings from CLIP show a scattered and overlapping distribution, indicating poor separation among classes, which makes accurate classification challenging. In contrast, as illustrated in Fig.\ref{fig:visualization}(b), our method effectively reorganizes the embeddings into more clearly defined clusters, significantly improving class separability and reducing feature overlap. To further clarify the advantage, we present a detailed visualization focusing on a random subset of 10 classes. Compared to CLIP's embeddings in Fig.\ref{fig:visualization}(c), our mean-shifted embeddings produce more compact and distinct clusters. Specifically, intra-class embeddings become notably tighter, and inter-class gaps are visibly enlarged, which reduces ambiguity at decision boundaries and facilitates more accurate predictions.

Another significant strength is the capability of our approach to overcome the inherent constraints of CLIP's original embedding space. As shown in Fig.\ref{fig:visualization}(e) and (f), we present a close-up comparison between classes 16 and 33. The original CLIP embeddings (Fig.\ref{fig:visualization}(e)) exhibit overlap, highlighting the difficulty in distinguishing closely related classes. However, after applying our mean-shift embedding technique (Fig.\ref{fig:visualization}(f)), the two classes become clearly separated with sharper decision boundaries. This confirms our hypothesis that leveraging local neighborhood information via mean-shift clustering effectively refines features, enhances discriminative capability, and thus overcomes intrinsic limitations of CLIP’s embedding space.

\subsection{Efficiency Analysis.}
We evaluate the efficiency of MS-TTA in terms of speed, memory, and accuracy. As shown in Table~\ref{tab:performance_comparison}, MS-TTA requires no training and runs at 10.05 FPS with only 1.4 GB memory. Despite its simplicity, MS-TTA achieves the best results in OOD benchmark performance and Cross-Dataset generalization, while being significantly faster than TPT (0.29 FPS) and DiffTPT (0.10 FPS).
These results demonstrate that MS-TTA offers a strong balance of accuracy and efficiency, making it well-suited for practical test-time adaptation.

\subsection{Plug-and-Play Adaptability of MS-TTA}
\label{sec:plug_and_play}

One of the key advantages of \textbf{MS-TTA} is its plug-and-play nature, allowing seamless integration into existing test-time adaptation (TTA) frameworks without modifying model architectures or requiring additional training. Unlike conventional adaptation techniques that rely on fine-tuning or hyperparameter-sensitive tuning, MS-TTA operates as a lightweight, training-free enhancement, refining feature representations solely through a single-step mean shift. This adaptability makes it a practical and scalable solution, particularly for real-world applications where retraining is computationally expensive or impractical due to dynamic data shifts.

To evaluate the compatibility of MS-TTA with existing methods, we apply it to two state-of-the-art training-free TTA approaches: TDA and BoostAdapter. 
{Importantly, when integrating with these methods, we keep its original positive/negative cache mechanism unchanged and only apply mean-shift refinement to the embeddings used for cache construction, retrieval, and logit aggregation.}
As shown in Table~\ref{tab:full_results_pp}, incorporating MS-TTA into these methods consistently improves classification accuracy across multiple datasets in the {Cross-Dataset Benchmark}. Specifically, when integrated with TDA, MS-TTA achieves an average performance gain of +1.24\%, with notable improvements on datasets such as EuroSAT (+3.93\%) and Flowers102 (+2.64\%). Similarly, adding MS-TTA to BoostAdapter results in an additional \textbf{+0.75\%} improvement, reinforcing the effectiveness of mean-shifted embeddings in refining decision boundaries. These results validate the compatibility of MS-TTA as a general enhancement module, demonstrating its ability to boost performance without requiring additional supervision or backpropagation.

Beyond its effectiveness, MS-TTA generalizes well across different datasets and domain shifts. As shown in Table~\ref{tab:full_results_pp}, it consistently improves adaptation on a diverse range of datasets, including structured domains such as Aircraft and EuroSAT and more complex, unstructured datasets like DTD and SUN397. The broad applicability of MS-TTA highlights its generalizability, making it a robust choice for training-free TTA across various real-world scenarios. Furthermore, since MS-TTA operates without modifying model weights or requiring iterative updates, it remains computationally efficient, enabling real-time adaptation with minimal overhead.

\section{Conclusion and Future Work}
\label{sec:Conclusion and Future Work}
We introduced MS-TTA, a training-free test-time adaptation framework that enhances feature representations beyond the original CLIP space using Mean-Shift clustering. Unlike prior methods that rely on high-confidence samples or pseudo-labels, MS-TTA refines all test samples, improving feature compactness and class separability. Extensive evaluations across OOD and Cross-Dataset Benchmarks confirm its effectiveness, consistently outperforming existing training-free approaches. Our method is efficient, requiring no additional training or model modifications, making it well-suited for real-world applications. Future work includes optimizing adaptive neighborhood selection and exploring broader applications across other vision-language models to enhance generalization.


\begin{thebibliography}{10}
\providecommand{\url}[1]{#1}
\csname url@samestyle\endcsname
\providecommand{\newblock}{\relax}
\providecommand{\bibinfo}[2]{#2}
\providecommand{\BIBentrySTDinterwordspacing}{\spaceskip=0pt\relax}
\providecommand{\BIBentryALTinterwordstretchfactor}{4}
\providecommand{\BIBentryALTinterwordspacing}{\spaceskip=\fontdimen2\font plus
\BIBentryALTinterwordstretchfactor\fontdimen3\font minus \fontdimen4\font\relax}
\providecommand{\BIBforeignlanguage}[2]{{%
\expandafter\ifx\csname l@#1\endcsname\relax
\typeout{** WARNING: IEEEtran.bst: No hyphenation pattern has been}%
\typeout{** loaded for the language `#1'. Using the pattern for}%
\typeout{** the default language instead.}%
\else
\language=\csname l@#1\endcsname
\fi
#2}}
\providecommand{\BIBdecl}{\relax}
\BIBdecl

\bibitem{clip}
A.~Radford, J.~W. Kim, C.~Hallacy \emph{et~al.}, ``Learning transferable visual models from natural language supervision,'' in \emph{Proceedings of the International Conference on Machine Learning}, 2021, pp. 8748--8763.

\bibitem{ALIGN}
C.~Jia, Y.~Yang, Y.~Xia \emph{et~al.}, ``Scaling up visual and vision-language representation learning with noisy text supervision,'' in \emph{Proceedings of the International Conference on Machine Learning}, 2021, pp. 4904--4916.

\bibitem{tpt}
M.~Shu, W.~Nie, D.A. Huang \emph{et~al.}, ``Test-time prompt tuning for zero-shot generalization in vision-language models,'' \emph{Advances in Neural Information Processing Systems}, vol.~35, pp. 14274--14289, 2022.

\bibitem{difftpt}
C.-M. Feng, K.~Yu, Y.~Liu, S.~Khan, and W.~Zuo, ``Diverse data augmentation with diffusions for effective test-time prompt tuning,'' in \emph{Proceedings of the IEEE/CVF International Conference on Computer Vision}, 2023, pp. 2704--2714.

\bibitem{Zhang2024Historical}
J.~Zhang, J.~Huang, X.~Zhang, L.~Shao, and S.~Lu, ``Historical test-time prompt tuning for vision foundation models,'' \emph{Advances in Neural Information Processing Systems}, vol.~37, pp. 12872--12896, 2024.

\bibitem{tda}
A.~Karmanov, D.~Guan, S.~Lu, A.~El~Saddik, and E.~Xing, ``Efficient test-time adaptation of vision-language models,'' in \emph{Proceedings of the IEEE/CVF Conference on Computer Vision and Pattern Recognition}, 2024, pp. 14162--14171.

\bibitem{Zhang2024BoostAdapter}
T.~Zhang, J.~Wang, H.~Guo, T.~Dai, B.~Chen, and S.-T. Xia, ``BoostAdapter: Improving vision-language test-time adaptation via regional bootstrapping,'' \emph{Advances in Neural Information Processing Systems}, vol.~37, pp. 67795--67825, 2024.

\bibitem{meanshift}
D.~Comaniciu and P.~Meer, ``Mean shift: A robust approach toward feature space analysis,'' \emph{IEEE Transactions on Pattern Analysis and Machine Intelligence}, vol.~24, no.~5, pp. 603--619, 2002.

\bibitem{cheng1995mean}
Y.~Cheng, ``Mean shift, mode seeking, and clustering,'' \emph{IEEE Transactions on Pattern Analysis and Machine Intelligence}, vol.~17, no.~8, pp. 790--799, 1995.

\bibitem{fukunaga1975estimation}
K.~Fukunaga and L.~Hostetler, ``The estimation of the gradient of a density function, with applications in pattern recognition,'' \emph{IEEE Transactions on Information Theory}, vol.~21, no.~1, pp. 32--40, 1975.

\bibitem{tta1}
M.~Boudiaf, R.~Mueller, I.~Ben~Ayed, and L.~Bertinetto, ``Parameter-free online test-time adaptation,'' in \emph{Proceedings of the IEEE/CVF Conference on Computer Vision and Pattern Recognition}, 2022, pp. 8344--8353.

\bibitem{memo}
M.~Zhang, S.~Levine, and C.~Finn, ``MEMO: Test time robustness via adaptation and augmentation,'' \emph{Advances in Neural Information Processing Systems}, vol.~35, pp. 38629--38642, 2022.

\bibitem{ptta}
L.~Yuan, B.~Xie, and S.~Li, ``Robust test-time adaptation in dynamic scenarios,'' in \emph{Proceedings of the IEEE/CVF Conference on Computer Vision and Pattern Recognition}, 2023, pp. 15922--15932.

\bibitem{domainadaptor}
J.~Zhang, L.~Qi, Y.~Shi, and Y.~Gao, ``DomainAdaptor: A novel approach to test-time adaptation,'' in \emph{Proceedings of the IEEE/CVF International Conference on Computer Vision}, 2023, pp. 18925-18935.

\bibitem{dota}
Z.~Han, J.~Yang \emph{et~al.}, ``DOTA: Distributional test-time adaptation of vision-language models,'' \emph{arXiv preprint arXiv:2409.19375}, 2024.

\bibitem{zhao2025expamoe}
J.~Zhao, C.~Ding, S.~Dong, J.~Li, Q.~Wang, Y.~He, and Y.~Gong, ``Shared \& Domain Self-Adaptive Experts with Frequency-Aware Discrimination for Continual Test-Time Adaptation,'' \emph{arXiv preprint arXiv:2507.00502}, 2025.


\bibitem{zenke2017continual}
F.~Zenke, B.~Poole, and S.~Ganguli, ``Continual learning through synaptic intelligence,'' in \emph{Proceedings of the 34th International Conference on Machine Learning (ICML)}, 2017, pp. 3987--3995.

\bibitem{han2025goal}
J.~Han, C.~Ding, S.~Dong \emph{et~al.},  ``GOAL: Geometrically Optimal Alignment for Continual Generalized Category Discovery,'' \emph{arXiv preprint arXiv:2602.19872}, 2026.

\bibitem{wang2025boostdil}
Q.~Wang, X.~Song, Y.~He \emph{et~al.}, ``Boosting domain incremental learning: Selecting the optimal parameters is all you need,'' in \emph{Proceedings of the IEEE/CVF Conference on Computer Vision and Pattern Recognition}, 2025, pp. 4839--4849.

\bibitem{han2025biag}
J.~Han, C.~Ding, Y.~He \emph{et~al.}, ``Learn by reasoning: Analogical weight generation for few-shot class-incremental learning,'' \emph{IEEE Transactions on Circuits and Systems for Video Technology}, 2025.


\bibitem{vaze2022gcd}
S.~Vaze, K.~Han, A.~Vedaldi, and A.~Zisserman, ``Generalized category discovery,'' in \emph{Proceedings of the IEEE/CVF Conference on Computer Vision and Pattern Recognition}, 2022, pp. 7492--7501.


\bibitem{wen2023parametric}
X.~Wen, B.~Zhao, and X.~Qi, ``Parametric classification for generalized category discovery: A baseline study,'' in \emph{Proceedings of the IEEE/CVF International Conference on Computer Vision}, 2023, pp.16544-16554.


\bibitem{han2025csua}
J.~Han, S.~Wang, Y.~He \emph{et~al.}, ``Consistent supervised-unsupervised alignment for generalized category discovery,'' \emph{arXiv preprint arXiv:2507.04725}, 2025.

\bibitem{augmix}
D.~Hendrycks, N.~Mu, E.~D. Cubuk \emph{et~al.}, ``Augmix: A simple data processing method to improve robustness and uncertainty,'' in \emph{International Conference on Learning Representations}, 2020.

\bibitem{stable}
R.~Rombach, A.~Blattmann, D.~Lorenz, P.~Esser, and B.~Ommer, ``High-resolution image synthesis with latent diffusion models,'' in \emph{Proceedings of the IEEE/CVF Conference on Computer Vision and Pattern Recognition}, 2022, pp. 10674-10685.

\bibitem{DPE}
C.~Zhang, S.~Stepputtis, K.~Sycara, and Y.~Xie, ``Dual prototype evolving for test-time generalization of vision-language models,'' \emph{Advances in Neural Information Processing Systems}, vol.~37, pp. 32111--32136, 2024.

\bibitem{zeng2025audio_tta}
R.~Zeng, Q.~Deng, R.~Zhang \emph{et~al.}, ``Exploring audio cues for enhanced test-time video model adaptation,'' \emph{IEEE Transactions on Circuits and Systems for Video Technology}, vol. 35, no. 12, pp. 12292-12305, 2025.

\bibitem{liu2024qed_tta}
J.~Liu, J.~Xie, F.~Zhou, and S.~He, ``Question type-aware debiasing for test-time visual question answering model adaptation,'' \emph{IEEE Transactions on Circuits and Systems for Video Technology}, vol.~34, no.~11, pp.~10805--10816, 2024.


\bibitem{wu2024ttagaze}
Y.~Wu, G.~Chen, L.~Ye \emph{et~al.}, ``TTAGaze: Self-supervised test-time adaptation for personalized gaze estimation,'' \emph{IEEE Transactions on Circuits and Systems for Video Technology}, vol.~34, no.~11, pp. 10959--10971, 2024.

\bibitem{chen2023cameraaware_tta}
K.~Chen, T.~Gong, and L.~Zhang, ``Camera-aware recurrent learning and earth mover's test-time adaption for generalizable person re-identification,'' \emph{IEEE Transactions on Circuits and Systems for Video Technology}, vol.~34, no.~1, pp.~357--370, 2024.


\bibitem{liu2025test}
J.~Liu and Z.~Yang, ``Test-time adaptation for real-world video adverse weather restoration with meta batch normalization,'' \emph{IEEE Transactions on Circuits and Systems for Video Technology}, vol.~35, no.~6, pp.~5533--5544, 2025.


\bibitem{FSTTA}
J.~Gao, X.~Yao, and C.~Xu, ``Fast-slow test-time adaptation for online vision-and-language navigation,'' in \emph{Proceedings of the International Conference on Machine Learning}, 2024, pp. 14902--14919.

\bibitem{tip}
R.~Zhang, W.~Zhang, R.~Fang \emph{et~al.}, ``Tip-adapter: Training-free adaption of clip for few-shot classification,'' in \emph{European Conference on Computer Vision}, 2022, pp. 493--510.

\bibitem{BCA}
L.~Zhou, M.~Ye, S.~Li \emph{et~al.}, ``Bayesian test-time adaptation for vision-language models,'' in \emph{Proceedings of the IEEE/CVF Conference on Computer Vision and Pattern Recognition}, 2025, pp. 29999--30009.

\bibitem{dong2025beyond}
S.~Dong, C.~Ding, J.~Li, J.~Han, Q.~Wang, Y.~He, and Y.~Gong, ``Beyond CLIP generalization: Against forward \& backward forgetting adapter for continual learning of vision-language models,'' \emph{arXiv preprint arXiv:2505.07690}, 2025.


\bibitem{downsteam1}
J.~Ding, N.~Xue, G.-S. Xia, and D.~Dai, ``Decoupling zero-shot semantic segmentation,'' in \emph{Proceedings of the IEEE/CVF Conference on Computer Vision and Pattern Recognition}, 2022, pp. 11583--11592.

\bibitem{downsteam2}
M.~Maaz, H.~Rasheed, S.~Khan, F.~S. Khan, R.~M. Anwer, and M.-H. Yang, ``Class-agnostic object detection with multi-modal transformer,'' in \emph{European Conference on Computer Vision}, 2022, pp. 512--531.

\bibitem{tan2025clip_cameraagnostic}
X.~Tan, X.~Gong, and Y.~Xiang, ``CLIP-based camera-agnostic feature learning for intra-camera supervised person re-identification,'' \emph{IEEE Transactions on Circuits and Systems for Video Technology}, vol.~35, no.~5, pp. 4100--4115, 2025.

\bibitem{cheng2022visualtext_pedestrian}
X.~Cheng, M.~Jia, Q.~Wang, and J.~Zhang, ``A simple visual-textual baseline for pedestrian attribute recognition,'' \emph{IEEE Transactions on Circuits and Systems for Video Technology}, vol.~32, no.~10, pp.~6994--7004, 2022.


\bibitem{fan2023parformer}
X.~Fan, Y.~Zhang, Y.~Lu, and H.~Wang, ``PARFormer: Transformer-based multi-task network for pedestrian attribute recognition,'' \emph{IEEE Transactions on Circuits and Systems for Video Technology}, vol.~34, no.~1, pp. 411--423, 2024.

\bibitem{comaniciu2000real}
D.~Comaniciu, V.~Ramesh, and P.~Meer, ``Real-time tracking of non-rigid objects using mean shift,'' in \emph{Proceedings of the IEEE Conference on Computer Vision and Pattern Recognition}, vol.~2, 2000, pp. 142--149.

\bibitem{kumar2022gridshift}
A.~Kumar, O.~S. Ajani, S.~Das, and R.~Mallipeddi, ``Gridshift: A faster mode-seeking algorithm for image segmentation and object tracking,'' in \emph{Proceedings of the IEEE/CVF Conference on Computer Vision and Pattern Recognition}, 2022, pp. 8121-8129.


\bibitem{jang2021meanshiftpp}
J.~Jang and H.~Jiang, ``MeanShift++: Extremely fast mode-seeking with applications to segmentation and object tracking,'' in \emph{Proceedings of the IEEE/CVF Conference on Computer Vision and Pattern Recognition}, 2021, pp. 4100-4111.



\bibitem{comaniciu2003kernel}
D.~Comaniciu, V.~Ramesh, and P.~Meer, ``Kernel-based object tracking,'' \emph{IEEE Transactions on Pattern Analysis and Machine Intelligence}, vol.~25, no.~5, pp. 564--577, 2003.

\bibitem{kong2018recurrent}
S.~Kong and C.~C. Fowlkes, ``Recurrent pixel embedding for instance grouping,'' in \emph{Proceedings of the IEEE Conference on Computer Vision and Pattern Recognition}, 2018, pp. 9018--9028.

\bibitem{koohpayegani2021mean}
S.~A. Koohpayegani, A.~Tejankar, and H.~Pirsiavash, ``Mean shift for self-supervised learning,'' in \emph{Proceedings of the IEEE/CVF International Conference on Computer Vision}, 2021, pp. 10326--10335.

\bibitem{fashing2005mean}
M.~Fashing and C.~Tomasi, ``Mean shift is a bound optimization,'' \emph{IEEE Transactions on Pattern Analysis and Machine Intelligence}, vol.~27, no.~3, pp. 471--474, 2005.

\bibitem{kobayashi2010hypersphere}
T.~Kobayashi and N.~Otsu, ``Von Mises-Fisher mean shift for clustering on a hypersphere,'' in \emph{Proceedings of the 20th International Conference on Pattern Recognition}, 2010, pp. 2130--2133.

\bibitem{chacon2019mixture}
J.~E. Chac{\'o}n, ``Mixture model modal clustering,'' \emph{Advances in Data Analysis and Classification}, vol.~13, no.~2, pp. 379--404, 2019.

\bibitem{yamasaki2023convergence}
R.~Yamasaki and T.~Tanaka, ``Convergence analysis of mean shift,'' \emph{IEEE Transactions on Pattern Analysis and Machine Intelligence}, vol.~46, no.~10, pp. 6688--6698, 2024.

\bibitem{li2007note}
X.~Li, Z.~Hu, and F.~Wu, ``A note on the convergence of the mean shift,'' \emph{Pattern Recognition}, vol.~40, no.~6, pp. 1756--1762, 2007.

\bibitem{singh2004robust}
M.~Singh, H.~Arora, and N.~Ahuja, ``A robust probabilistic estimation framework for parametric image models,'' in \emph{European Conference on Computer Vision}, 2004, pp. 508--522.

\bibitem{anand2013semi}
S.~Anand, S.~Mittal, O.~Tuzel, and P.~Meer, ``Semi-supervised kernel mean shift clustering,'' \emph{IEEE Transactions on Pattern Analysis and Machine Intelligence}, vol.~36, no.~6, pp. 1201--1215, 2014.

\bibitem{yuan2010agglomerative}
X.-T. Yuan, B.-G. Hu, and R.~He, ``Agglomerative mean-shift clustering,'' \emph{IEEE Transactions on Knowledge and Data Engineering}, vol.~24, no.~2, pp. 209--219, 2012.

\bibitem{imagenet}
J.~Deng, W.~Dong, R.~Socher, L.-J. Li, K.~Li, and L.~Fei-Fei, ``Imagenet: A large-scale hierarchical image database,'' in \emph{Proceedings of the IEEE Conference on Computer Vision and Pattern Recognition}, 2009, pp. 248--255.

\bibitem{imageneta}
D.~Hendrycks, K.~Zhao, S.~Basart, J.~Steinhardt, and D.~Song, ``Natural adversarial examples,'' in \emph{Proceedings of the IEEE/CVF Conference on Computer Vision and Pattern Recognition}, 2021, pp. 15262--15271.

\bibitem{imagenetr}
D.~Hendrycks, S.~Basart, N.~Mu \emph{et~al.}, ``The many faces of robustness: A critical analysis of out-of-distribution generalization,'' in \emph{Proceedings of the IEEE/CVF International Conference on Computer Vision}, 2021, pp.~8320--8329.

\bibitem{imagenetv2}
B.~Recht, R.~Roelofs, L.~Schmidt, and V.~Shankar, ``Do ImageNet classifiers generalize to ImageNet?'' in \emph{Proceedings of the International Conference on Machine Learning}, 2019, pp. 5389--5400.

\bibitem{imagenetsk}
H.~Wang, S.~Ge, E.~P.~Xing, and Z.C.~Lipton, ``Learning robust global representations by penalizing local predictive power,'' in \emph{Advances in Neural Information Processing Systems}, vol.~32, pp.~10506--10518, 2019.


\bibitem{caltech101}
Li~Fei-Fei, R.~Fergus, and P.~Perona, ``Learning generative visual models from few training examples: An incremental Bayesian approach tested on 101 object categories,'' in \emph{Proceedings of the 2004 Conference on Computer Vision and Pattern Recognition Workshop}, 2004, p. 178.


\bibitem{pets}
O.~M. Parkhi, A.~Vedaldi, A.~Zisserman, and C.~V. Jawahar, ``Cats and dogs,'' in \emph{Proceedings of the IEEE Conference on Computer Vision and Pattern Recognition}, 2012, pp. 3498--3505.

\bibitem{cars}
J.~Krause, M.~Stark, J.~Deng, and L.~Fei-Fei, ``3D object representations for fine-grained categorization,'' in \emph{Proceedings of the IEEE International Conference on Computer Vision Workshops}, 2013, pp. 554--561.

\bibitem{flowers}
M.-E. Nilsback and A.~Zisserman, ``Automated flower classification over a large number of classes,'' in \emph{Proceedings of the Sixth Indian Conference on Computer Vision, Graphics and Image Processing}, 2008, pp. 722--729.

\bibitem{food101}
L.~Bossard, M.~Guillaumin, and L.~Van~Gool, ``Food-101: Mining discriminative components with random forests,'' in \emph{European Conference on Computer Vision}, 2014, pp. 446--461.

\bibitem{aircraft}
S.~Maji, E.~Rahtu, J.~Kannala, M.~Blaschko, and A.~Vedaldi, ``Fine-grained visual classification of aircraft,'' \emph{arXiv preprint arXiv:1306.5151}, 2013.

\bibitem{sun397}
J.~Xiao, J.~Hays, K.~A. Ehinger, A.~Oliva, and A.~Torralba, ``SUN database: Large-scale scene recognition from abbey to zoo,'' in \emph{Proceedings of the IEEE Computer Society Conference on Computer Vision and Pattern Recognition}, 2010, pp. 3485--3492.

\bibitem{dtd}
M.~Cimpoi, S.~Maji, I.~Kokkinos, S.~Mohamed, and A.~Vedaldi, ``Describing textures in the wild,'' in \emph{Proceedings of the IEEE Conference on Computer Vision and Pattern Recognition}, 2014, pp. 3606--3613.

\bibitem{eurosat}
P.~Helber, B.~Bischke, A.~Dengel, and D.~Borth, ``EuroSAT: A novel dataset and deep learning benchmark for land use and land cover classification,'' \emph{IEEE Journal of Selected Topics in Applied Earth Observations and Remote Sensing}, vol.~12, no.~7, pp. 2217--2226, 2019.

\bibitem{ucf101}
K.~Soomro, A.~R.~Zamir, and M.~Shah, ``Ucf101: A dataset of 101 human actions classes from videos in the wild,'' arXiv preprint arXiv:1212.0402, 2012.


\bibitem{coop}
K.~Zhou, J.~Yang, C.~C. Loy, and Z.~Liu, ``Learning to prompt for vision-language models,'' \emph{International Journal of Computer Vision}, vol.~130, no.~9, pp. 2337--2348, 2022.

\bibitem{cocoop}
K.~Zhou, J.~Yang, C.~C. Loy, and Z.~Liu, ``Conditional prompt learning for vision-language models,'' in \emph{Proceedings of the IEEE/CVF Conference on Computer Vision and Pattern Recognition}, 2022, pp. 16816--16825.

\bibitem{mta}
M.~Zanella and I.~Ben~Ayed, ``On the test-time zero-shot generalization of vision-language models: Do we really need prompt learning?'' in \emph{Proceedings of the IEEE/CVF Conference on Computer Vision and Pattern Recognition}, 2024, pp. 23783--23793.

\bibitem{attention_is_all_you_need}
A.~Vaswani, N.~Shazeer, N.~Parmar, \emph{et al.}, ``Attention is all you need,'' in \emph{Advances in Neural Information Processing Systems}, vol.~30, pp.~5998--6008, 2017.


\bibitem{resnet}
K.~He, X.~Zhang, S.~Ren, and J.~Sun, ``Deep residual learning for image recognition,'' in \emph{Proceedings of the IEEE Conference on Computer Vision and Pattern Recognition}, 2016, pp. 770--778.

\bibitem{vit}
A.~Dosovitskiy, \emph{et al.}, ``An image is worth 16$\times$16 words: Transformers for image recognition at scale,'' \emph{arXiv preprint arXiv:2010.11929}, 2020.

\end{thebibliography}

\end{document}